\documentclass{article}

\usepackage[final]{neurips_2024}

\usepackage[utf8]{inputenc} % allow utf-8 input
\usepackage[T1]{fontenc}    % use 8-bit T1 fonts
\usepackage{hyperref}       % hyperlinks
\usepackage{url}            % simple URL typesetting
\usepackage{booktabs}       % professional-quality tables
\usepackage{amsfonts}       % blackboard math symbols
\usepackage{nicefrac}       % compact symbols for 1/2, etc.
\usepackage{microtype}      % microtypography
\usepackage{xcolor}         % colors

\usepackage{cleveref}      % clever referencing (must be after hyperref)

\usepackage[textsize=tiny]{todonotes}
\usepackage{wrapfig}

\usepackage{algorithm}
\usepackage{algpseudocode}

\usepackage[textsize=tiny]{todonotes}
\usepackage{xspace}
\usepackage{bbm}
\usepackage{subcaption}
\usepackage{wrapfig}
\usepackage{multirow}
\usepackage{xcolor}
\usepackage{graphicx}
\usepackage[skip=5pt,font=small]{caption}

\title{REDUCR: Robust Data Downsampling using Class Priority Reweighting}

\author{%
  William Bankes\\ 
  Department of Computer Science\\
  University College London\\
  \texttt{william.bankes.21@ucl.ac.uk} \\
  \And
  George Hughes \\
  Department of Computer Science \\
  University College London \\
  \AND
  Ilija Bogunovic\thanks{Co-senior authors. Code available at: \href{https://github.com/williambankes/REDUCR}{https://github.com/williambankes/REDUCR}.} \\
  Department of Electrical Engineering \\
  University College London \\
  \texttt{i.bogunovic@ucl.ac.uk} \\
  \And
  Zi Wang$^*$ \\
  Google DeepMind \\
  \texttt{wangzi@google.com} \\
}

\begin{document}

\maketitle

\begin{abstract}
  Modern machine learning models are becoming increasingly expensive to train for real-world image and text classification tasks, where massive web-scale data is collected in a streaming fashion. To reduce the training cost, online batch selection techniques have been developed to choose the most informative datapoints. However, many existing techniques are not robust to class imbalance and distributional shifts, and can suffer from poor worst-class generalization performance. This work introduces \algnamens, a robust and efficient data downsampling method that uses class priority reweighting. \algname \emph{reduces} the training data while preserving worst-class generalization performance. \algnamens assigns priority weights to datapoints in a class-aware manner using an online learning algorithm. We demonstrate the data efficiency and robust performance of \algnamens on vision and text classification tasks. On web-scraped datasets with imbalanced class distributions, \algnamens significantly improves worst-class test accuracy (and average accuracy), surpassing state-of-the-art methods by around 15\%.
\end{abstract}

\section{Introduction}
\label{sec:Introduction}

The abundance of data has had a profound impact on machine learning (ML), both positive and negative. On the one hand, it has enabled ML models to achieve unprecedented performance on a wide range of tasks, such as image and text classification~\citep{kuznetsova2020open,he2015delving,brown2020language,tran2022plex,anil2023palm}. On the other hand, training models on such large datasets can demand significant computational resources~\citep{kaddour2023challenges}, making it unsustainable in some situations~\citep{bender2021dangers,patterson2021carbon}. Additionally, the high speed at which streaming data is collected can make it infeasible to train on all of the data before deployment. To tackle these issues, various methods have emerged to selectively choose training data, either through pre-training data pruning ~\citep{sorscher2022beyond, bachem2017practical} or online batch selection techniques ~\citep{loshchilov2015online, mindermann2022prioritized}, ultimately reducing data requirements and enabling ML models to handle otherwise unmanageable large and complex datasets.\looseness=-1

In real-world settings, a variety of factors can affect the selection of datapoints, such as noise~\citep{xiao2015learning, cao2020heteroskedastic, wei2021robust} and class-imbalance in the data~\citep{van2018inaturalist, philip1998toward, radivojac2004classification}. Online selection methods can exacerbate these problems by further reducing the number of datapoints from underrepresented classes, which can degrade the performance of the model on those classes~\citep{buda2018systematic, cui2019class}. Moreover, distributional shift~\citep{koh2021wilds} between training and test time can lead to increased generalization error if classes with poor generalization error are overrepresented at test time. 

\begin{figure*}[t!]
    \vspace{-5pt}
    \centering
    \includegraphics[width=.9\textwidth]{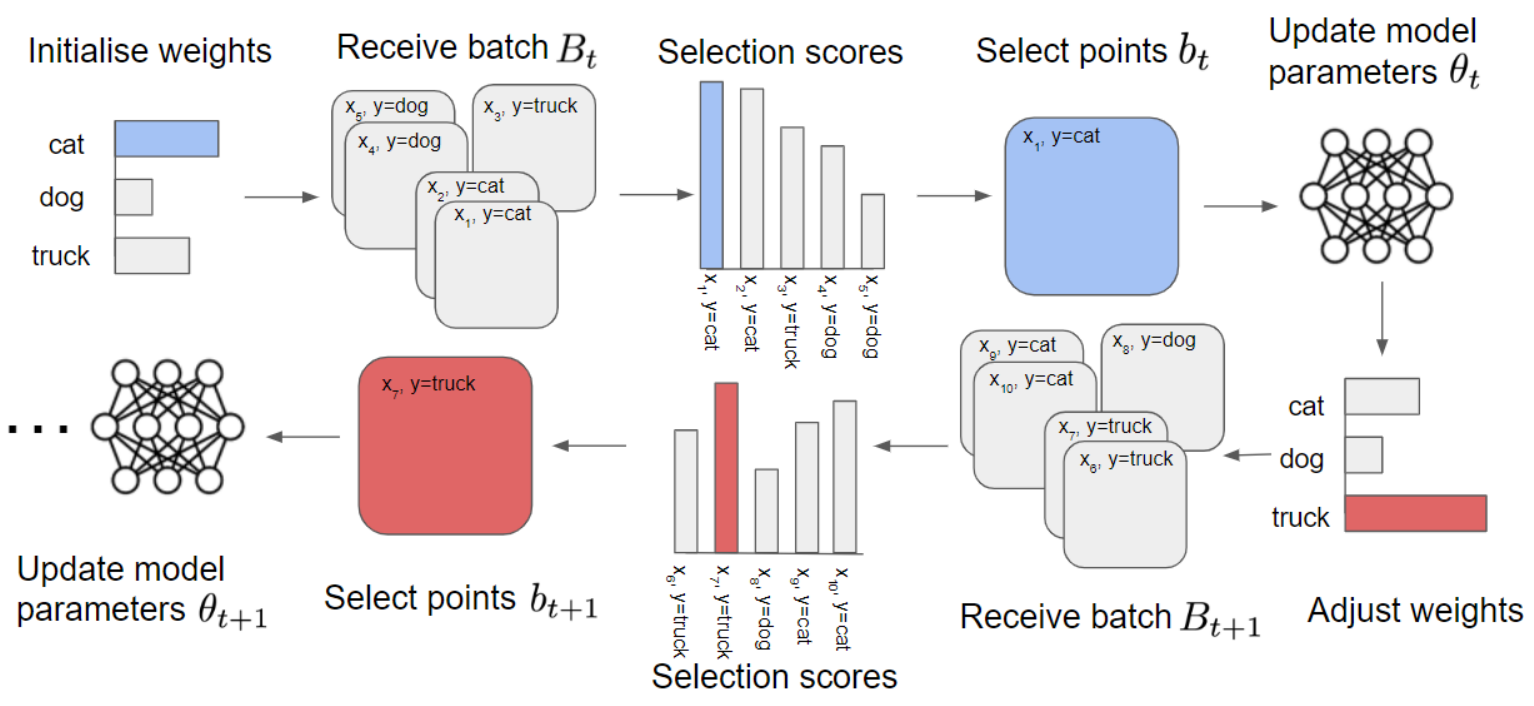}
    \vspace{-5pt}
    \caption{\algname starts by initializing weights of classes. At each timestep $t$, the model receives a batch of datapoints $B_t$. \algname computes the selection scores for each datapoint based on its usefulness to the model and the class weights, and selects new datapoints $b_t\subset B_t$ that achieve the highest selection scores. After the model takes gradient steps on the selected datapoints, \algname adjusts the weights to reflect increased priorities on underperforming classes.} 

    \label{fig:robust online batch selection protocol}
    \vspace{-17pt}
\end{figure*}

% \begin{wrapfigure}{R}{0.45\textwidth}
% \vspace{-10pt}
%     \centering  \includegraphics[width=0.4\textwidth]{Clothing1M_checkpointed_results.pdf}
%     \vspace{-10pt}
%      \caption{\algname significantly improves worst-class test accuracy on Clothing1M. \looseness=-1}
%       \label{fig:clothing1m worst class}
% \end{wrapfigure}

% \begin{figure}
%     \centering
%     \includegraphics[width=0.35\textwidth]{Clothing1M_checkpointed_results.pdf}
%      \caption{\algname significantly improves worst-class test accuracy on Clothing1M. \looseness=-1}
%       \label{fig:clothing1m worst class}
%       \vspace{-20pt}
% \end{figure}

\begin{wrapfigure}{R}{0.45\textwidth}
    \vspace{-0.5em}
    \centering  \includegraphics[width=0.4\textwidth]{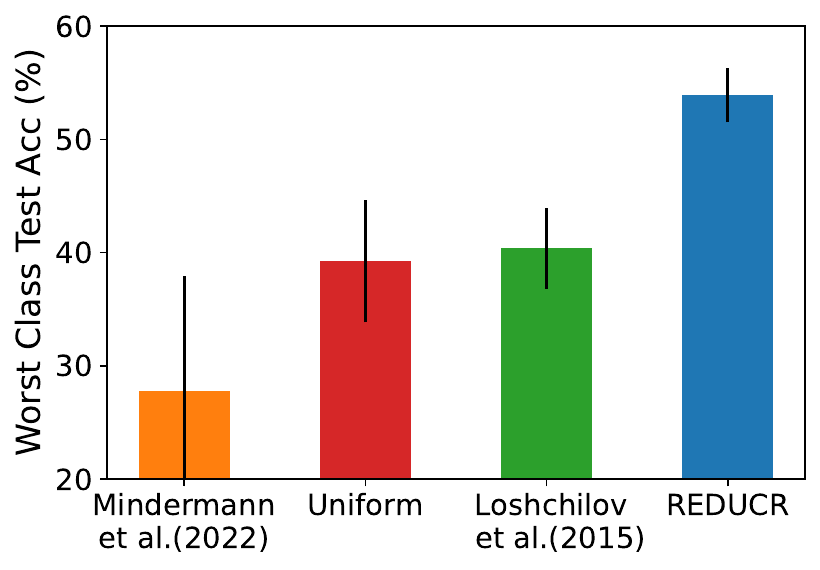}
     \caption{\algname significantly improves worst-class test accuracy on Clothing1M outperforming Uniform and other recent works.\looseness=-1} 
       \label{fig:clothing1m worst class}
    \vspace{-2em}
\end{wrapfigure}

In this work, we introduce \algname, which is a new online batch selection method that is \emph{robust} to noise, imbalance, and distributional shifts. \algname employs multiplicative weights update to reweight and prioritize classes that are performing poorly during online batch selection. \Cref{fig:robust online batch selection protocol} illustrates the intuition behind how the method works. \algname can effectively reduce the training data while preserving the worst-class generalization performance of the model. For example, on the Clothing1M dataset~\citep{xiao2015learning}, \Cref{fig:clothing1m worst class} shows that, compared to the best performing online batch selection methods, \algname achieves around a 15\% boost in performance for the worst-class test accuracy.
\looseness=-1

%\vspace{-1em}
\paragraph{Main contributions.} (1) We formalise the maximin problem of robust data downsampling (\S\ref{ssec:problem_formulation}). (2) We propose the \algname algorithm, which is equipped with a new robust selection rule that evaluates how much datapoints will affect the generalization error of a specific class (\S\ref{sec:robust_selection_rule}). (3) We evaluate our algorithm on a series of text and image classification tasks and show that it achieves strong worst-class test accuracy while frequently surpassing state-of-the-art methods in
terms of average test accuracy(\S\ref{sec:experiments}).\looseness=-1

%\vspace{-1em}
\paragraph{Related work.} 
%\section{Related Work}
\citet{mindermann2022prioritized} have developed an online batch selection method called \textsc{RHO-Loss}, which uses a \emph{reference model} trained on a holdout dataset to guide the selection of points during training. Certain extensions of this work have focused on using a reference model in different settings such as reinforcement learning~\citep{sujit2022prioritizing}. However, to our knowledge, none have focused on improving the worst-class generalisation performance. Other batch selection methods \citep{loshchilov2015online, jiang2019accelerating, kawaguchi2020ordered} use the training loss of points under the model or an approximate gradient norm~\citep{katharopoulos2017biased} to select challenging points. We observe that these methods (e.g., see \citet{loshchilov2015online} in \Cref{fig:clothing1m worst class}) exhibit greater consistency in terms of worst-class generalization error in imbalanced datasets. Nevertheless, \citet{loshchilov2015online} do not surpass the average generalization error achieved by point selection with a reference model, namely, \textsc{RHO-Loss}. \looseness=-1   
Recently, several works have also used reference models or a holdout dataset to train robust models. \citet{oren2019distributionally, liu2021just, clark2019don} use a reference model to identify difficult-to-learn groups (or points, or biases) during training. \citet{han2018co} use two models which act as a reference model for the other to remove noisy points from the training data. \citet{cao2020heteroskedastic,ren2018learning} use a holdout dataset to reweight points or their regularization during training to achieve the best loss on the validation holdout dataset. 

\citet{sagawa2019distributionally} reweight groups known at training time and focus on fighting spurious correlations and improving worst-group generalisation error. In contrast, in our setting, class labels are available and we measure the performance in terms of worst-class generalisation error. Moreover, whilst these works aim to train robust models they do not consider efficient data downsampling strategies. The approach of \citet{sagawa2019distributionally} to group robustness considers a small number of groups (up to 4 in their empirical study). Similarly, in our work, we consider classification settings with a controlled number of classes (< 1000) as the problem of robustness becomes less applicable in settings where the number of classes are high.\looseness=-1

%and the performance scales linearly ($\mathcal{\Tilde{O}}(G)$) in G, the number of groups, as such group robustness methods do not consider settings in which the number of groups grows too large.

\citet{xie2023doremi} use both weights update rules and a reference model to find mixtures of corpora in LLM pretraining resulting in improved performance and training speed. Besides the problem setup, our method differs in three ways: i) we focus upon online batch selection; ii) we use multiple reference models; iii) and we use a class-holdout loss term (see \Cref{eq:robust_rho_loss_restated}) to reweight batches. Efficient data downsampling is a well-explored problem with various approaches, including active learning methods when label information is unknown~\citep{mackay1992information, houlsby2011bayesian, kirsch2019batchbald, kirsch2021stochastic, ash2019deep}; data pruning and coreset techniques for pre-training data downsampling~\citep{sorscher2022beyond, bachem2017practical, borsos2020coresets, coleman2019selection}; data distillation approaches~\citep{cazenavette2022dataset, nguyen2021dataset}; and non-parametric inducing point methods~\citep{galy2021adaptive}.\looseness=-1
\vspace{-2pt}
\section{Background}
\vspace{-2pt}
\label{sec:prelims}

We consider a $C$-way classification task and denote a model as $p(y \mid x, \theta)$, where $x$ denotes an input and $y\in[C]$ the corresponding class label; the model is parameterized by $\theta$. For any training dataset $\mathcal{D} = \{(x_i, y_i)\}_{i=1}^{N}$ with $N$ datapoints, we use a point estimate of $\theta$ to approximate the posterior model as $p(y\mid x,\mathcal{D}) \approx p(y\mid x,\hat \theta)$. This estimate $\hat \theta$ can be obtained by running stochastic gradient descent (SGD) to optimize the cross-entropy loss over a training dataset  $\mathcal{D}$.\looseness=-1

\vspace{0.3em}

\label{sec:online_batch_selection_subsection}
The goal of \emph{data downsampling} is to select a dataset $\DT \subset \mathcal{D}$ of size $T$ ($\ll N$) for training such that the generalisation error of the resulting model is minimised. We write this objective in terms of a separate \emph{holdout} dataset $\Dh = \{(x_{ho,i}, y_{ho,i})\}_{i=1}^{N_{ho}}$ as follows:\looseness=-1 
\begin{equation}
    \label{eq:batch_selection}
    \DT = \argmaxDT \, \log p(\mathbf{y}_{ho}|\mathbf{x}_{ho},D),
\end{equation}
where the inputs and their labels are $\xho = [x_{i,ho} ]_{i=1}^{N_{ho}}$ and $\yho = [y_{i,ho}]_{i=1}^{N_{ho}}$, respectively. Here, the likelihood of the holdout dataset is used as a proxy for the generalisation error. The problem is computationally prohibitive due to its combinatorial nature. Moreover, for a massive (or streaming) training dataset $\mathcal D$, it is not computationally possible to load $\mathcal D$ all at once and it is common to loop through the data by iteratively loading subsets. \looseness=-1 

\textbf{Online batch selection} is a practical streaming setup to approximate the data downsampling problem, where at each timestep $t$, the model observes a training data subset $B_t \subset \mathcal{D}$, and the goal is to iteratively select a small batch $b_t\subset B_t$ for the model to take gradient steps. A standard solution to this problem is to design a selection score function that take into account the labels of the data.  The selection score function can then be used to \emph{score} the utility of the small batch $b_t$. See \Cref{alg:online batch selection} in \Cref{sec:online batch selection pseudo code} for an example method.

\vspace{0.1em}
\textbf{Reducible Holdout Loss (RHO-Loss)}~\citep{mindermann2022prioritized} is an online batch selection method that uses the performance on a holdout dataset as the selection scores for small batches. More precisely, for each timestep $t$, RHO-Loss selects  
\begin{equation}
    \label{eq:rho_loss}
    b_t = \underset{b \subset B_t}{\mathrm{argmax}}\; \log p(\yho\mid\xho, \mathcal{D}_{t} \cup b),  
\end{equation}
where $\mathcal{D}_{t} = \bigcup_{\tau = 1}^{t-1} b_{\tau}$ is the cumulative training data the model has encountered until iteration $t$. 

\vspace{-2pt}
\section{Problem Formulation}
\vspace{-2pt}
\label{ssec:problem_formulation}
In this work, we introduce the \emph{robust} data downsampling problem, where the goal is to select a training dataset $\DT$ of size $T$ such that worst-class performance is optimized. Let the holdout dataset with class $c\in[C]$ be $D_{ho}^{(c)} = \{(x,y)\in  D_{ho}\mid y\equiv c\} = \{(x_{ho,i}^{(c)}, y_{ho,i}^{(c)})\}_{i=1}^{N_{ho}^{(c)}}$.  We can write the objective of robust data downsampling as  
\begin{equation}\label{eq:robust_batch_selection}
    \DT = \argmaxDT \min_{c \in [C]} \, \log p(\mathbf{y}_{ho}^{(c)}\mid\mathbf{x}_{ho}^{(c)}, D),
\end{equation}
where $\xhoc = [x_{ho,i}^{(c)}]_{i=1}^{N_{ho}^{(c)}}$ and $\yhoc = [y_{ho,i}^{(c)}]_{i=1}^{N_{ho}^{(c)}}$ correspond to the collections of inputs and labels in the class-specific holdout dataset $D_{ho}^{(c)}$.

Compared to \Cref{eq:batch_selection}, the objective in \Cref{eq:robust_batch_selection} is even more challenging because of the maximin optimisation that involves $C$ discrete classes. In fact, solving \Cref{eq:robust_batch_selection} is known to be NP-hard, even when the objectives (each $p(\mathbf{y}_{ho}^{(c)}|\mathbf{x}_{ho}^{(c)}, \cdot)$, $c \in [C]$) are \emph{submodular} set functions. \citet{chen2017robust} demonstrate the application of zero-sum game no-regret dynamics, where a learner employs a $(1-1/e)$-near-optimal greedy strategy and an adversary seeks to find a distribution over loss functions that maximizes the learner's loss. In this scenario, a single set is identified, which, although larger than size $T$, achieves a constant-factor approximation.

\textbf{Robust online batch selection} approximates the robust data downsampling problem by taking into account the practical limitations of data operation. Namely, we assume a streaming setting where the model observes training data subset $B_t\subset \mathcal D$ at each timestep $t$. The goal is to select a small batch $b_t\subset B_t$ to compute gradients for model training with SGD, such that the model obtains top performance for the worst-class (\Cref{eq:robust_batch_selection}). 
The robust setting motivates the development of novel batch selection methods that consider how each datapoint affects the generalization error on the worst-case class of inputs, rather than just the overall generalization error. Next, we introduce a new selection rule that achieves this and propose a practical algorithm for its implementation.

\vspace{-2pt}
\section{REDUCR for Robust Online Batch Selection}
\vspace{-2pt}
\label{section:robust_obs}
 
%% ROBUST ONLINE BATCH SELECTION INTUITION

We propose REDUCR, a \underline{r}obust and \underline{e}fficient data  \underline{d}ownsampling method  \underline{u}sing  \underline{c}lass priority \underline{r}eweighting to solve the robust online batch selection problem in \Cref{ssec:problem_formulation}. The batch selection strategy of REDUCR relates the effect of training on a batch of candidate points $b_t$ to the generalization error of a specific class in the holdout dataset.
\vspace{-2pt}
\subsection{Online Learning}
\vspace{-2pt}
\label{sec:online algorithm}
To solve \Cref{eq:robust_batch_selection} in an online manner, we propose to use class priority reweighting, a variant of the multiplicative weights update method \citep{freund1997decision,cesa2006prediction, sessa2019noregret}.  
At the beginning of training we initialise a weight vector $\mathbf{w}_0$ over a $C$ dimensional simplex, $ \Delta = \{ \mathbf{w}=[w_c]_{c=1}^C \in \mathbb{R}^{C} | \sum_{c = 1}^{C} w_{c} = 1\}$. Each element of $\mathbf{w}_0$ is initialised to be $w_{0, c} = 1/C$. For each iteration $t$, small batch $b_t\subset B_t$ is chosen by maximising the weighted sum of the $C$ different class-specific scoring functions (i.e., by best-responding to the current class-weights $\mathbf{w}_t$),
\vspace{-1em}
\begin{equation}
\label{eq:online weighted selection rule}
    b_t = \underset{b \subset B_t}{\mathrm{argmax}}\sum_{c=1}^C w_{t,c}\left( \log p(\yhoc|\xhoc, \mathcal{D}_{t} \cup b)\right),
\end{equation}
where $\mathcal D_t = \bigcup_{\tau = 1}^{t-1} b_{\tau}$, $\mathbf{w}_t = [w_{t,c}]_{c=1}^{C} \in \Delta$, and
\begin{equation}
\label{eq:weight update}
    w_{t,c} = w_{t-1,c} \tfrac{\exp \left(-\eta \log p(\yhoc|\xhoc, \mathcal{D}_{t})\right)}{\sum_{j=1}^C w_{t-1,j} \exp \left( -\eta \log p(\mathbf{y}_{ho}^{(j)}|\mathbf{x}_{ho}^{(j)}, \mathcal{D}_{t}) \right)}.
\end{equation}
In the previous alternating procedure, class-weights are updated multiplicatively according to how well they perform given the selected batch, they increase for poorly performing classes and decrease otherwise. In \Cref{eq:weight update}, $\eta$ is a learning rate that adjusts how concentrated the probability mass is in the resulting distribution. \Cref{fig:robust online batch selection protocol} shows an intuitive illustration of how reweighting works in practice where classes that perform badly have low data likelihoods and are thus upweighted by \Cref{eq:weight update}. In \Cref{app:payoff} we explore an alternative solution to solve \Cref{eq:robust_batch_selection}; we solve an approximate robust optimisation problem directly at every timestep $t$ and empirically demonstrate the multiplicative weights method outperforms it. We next introduce how to compute the likelihoods for class-specific holdout sets, i.e., $p(\yhoc|\xhoc, \mathcal{D}_{t} \cup b)$ in \Cref{eq:online weighted selection rule}.\looseness=-1
\vspace{-2pt}
\subsection{Computing selection scores} 
\vspace{-2pt}
\label{sec:robust_selection_rule}
Given the current dataset $\Dt$ at timestep $t$ and additional datapoints $b\subset B_t$, we would like to compute the likelihood of the holdout dataset that belongs to class $c$.
 For simplicity, we consider the case where the small batch to be selected only includes a single datapoint, i.e., $b = \{(x,y)\}$. 
We express the objective using a Bayesian perspective,    
\begin{align}
\log p(\yhoc\,|\, \xhoc, \mathcal{D}_{t} \cup \{(x, y)\})
&= \log \tfrac{p(y\mid x,\Dhoc, \Dt) p(\yhoc \mid \xhoc, x, \Dt)}{p(y\mid x, \xhoc, \Dt)} \label{eq:Bayes_rule}\\ 
&= \log \tfrac{p(y\mid x, \mathcal{D}_{ho}^{(c)}, \mathcal{D}_{t})p(\mathbf{y}_{ho}^{(c)} \mid \mathbf{x}_{ho}^{(c)},\mathcal{D}_{t})}{p(y \mid x,\mathcal{D}_{t})} \\
&= - \log p(y\,|\, x, \mathcal{D}_{t}) + \log p(y\,|\, x, \mathcal{D}_{t}, \mathcal{D}_{ho}^{(c)}) + \log p(\mathbf{y}_{ho}^{(c)}\,|\,\mathbf{x}_{ho}^{(c)}, \mathcal{D}_{t}) \nonumber.
\end{align}
\Cref{eq:Bayes_rule} follows from the application of the Bayes rule and the conditional independence of $x$ and $\xhoc$ with $\yhoc$ and $y$, respectively. The posterior terms in \Cref{eq:Bayes_rule} can be approximated with point estimates of model parameters (see \S\ref{sec:prelims}). Computing \Cref{eq:Bayes_rule} involves two models: (1) the \emph{target} model with parameters $\theta_t$, which is trained on the cumulative training dataset $\mathcal D_t = \bigcup_{\tau = 1}^{t-1} b_{\tau}$; (2) a \emph{class-irreducible loss model} (following the terminology from \citet{mindermann2022prioritized}) with parameters $\theta_t^{(c)}$, which is trained on $\mathcal D_t$ and class-specific holdout data $\mathcal D_{ho}^{(c)}$. The target model is what we are interested in for the classification task. We use $\mathcal L[y| x, \theta] = -\log p(y\,|\, x, \theta)$ to denote the cross-entropy loss for any model parameters $\theta$, and we re-write \Cref{eq:Bayes_rule} as follows,
\begin{equation}
          \log p(\yhoc\mid \xhoc, \mathcal{D}_{t} \cup \{(x, y)\}) \approx\; \underbrace{\targetl}_{\text{model loss}} - \underbrace{\irredlc}_{\text{class-irreducible loss}} - \underbrace{\robusthl}_{\text{class-holdout loss}}.
\label{eq:robust_rho_loss_restated}
\end{equation}
We name the three terms in \Cref{eq:robust_rho_loss_restated} the \emph{model loss}, \emph{class-irreducible loss} and \emph{class-holdout loss}, respectively. We define the term \emph{excess loss} as the difference of the model loss and class-irreducible loss. The excess loss is the improvement in loss for point $(x,y)$ by observing more data from class $c$ (i.e., $\Dhoc$). Intuitively, if two data points are from different classes, \algname will take into account the weight of the worst-performing class, which is reflected by the class-holdout loss. This ensures that \algname is focusing on improving the performance of the model on the classes that are most difficult to learn. In a different scenario, if two datapoints are from the same class, their class-holdout losses will be the same, and the point with a larger excess loss will be preferred. This means that \algname prefers datapoints whose losses have more potential to be improved. 

Computing the approximate in \Cref{eq:robust_rho_loss_restated} is far more tractable than naively re-training a new model (i.e.,  $\log p(\yhoc|\xhoc, \mathcal{D}_{t} \cup \{(x, y)\})$) for each possible candidate point $(x,y)$. 
The model loss and the class-holdout loss only require evaluating the cross-entropy losses of some datapoints on the target model. More generaly, if batch $b$ can include more than one point, we can simply change the $x$ and $y$ to a list of inputs and labels instead. Next, we further improve the efficiency of REDUCR by approximating the class-irreducible loss model.

%\vspace{-2pt}
\subsection{Class-Irreducible Loss Models} 
%\vspace{-2pt}
\label{sec:class irred loss model}

For each selected batch $b_t$ under the current selection rule in \Cref{eq:robust_rho_loss_restated}, we need to update $C$ class-irreducible loss models to compute the class-irreducible losses. We propose to approximate these models using \emph{amortised} class-irreducible loss models, which are trained for each class at the beginning of REDUCR and do not need to be updated during online batch selection.\looseness=-1 

We interpret the class irreducible loss term as an expert model at predicting the label of points from a specific class $c$ due to the extra data from the holdout dataset this term has available. To create an approximation of this expert model, we train the amortised class-irreducible loss models using an adjusted loss function in which points with a label from the class $c$ are up-weighted by a parameter $\gamma \in (0,+\infty)$ (set in Section~\ref{sec:experiments}): 
\begin{equation}\label{eq:amortised}
  \phi_c = \argmin_{\phi} \sum_{(x,y) \in \mathcal{D_{ho}}} (1 + \gamma\,\mathbb{I}[c \equiv y]) \,\mathcal{L}[y|x,\phi].
\end{equation}
Here we define $\mathbb{I}[\cdot]$ as the indicator function. \Cref{eq:amortised} optimizes over the parameters of the amortised class-irreducible loss model for class $c$, and obtain $\phi_c$ to approximate $\theta_t^{(c)}$ in \Cref{eq:robust_rho_loss_restated}, i.e.,
$\irredlc \approx \mathcal{L}[y|x, \phi_c].$ The up-weighting of points can be considered a form of importance weighting \citep{shimodaira2000improving}, where by up-weighting points with labels in a specific class we calculate a Monte Carlo approximation of the loss under a distribution in which points from class $c$ are more prevalent. \Cref{alg:class irred model training} details the full amortised class-irreducible loss model training procedure in \Cref{app:class irred model training}. We provide further motivation of our approximation in \Cref{app:class irred model approx}.\looseness=-1

\vspace{-10pt}
\subsection{REDUCR as a practical algorithm}
%\vspace{-3pt}
We use the selection objective in \Cref{eq:robust_rho_loss_restated} along with the amortised class-irreducible loss model approximation (\Cref{sec:class irred loss model}) and the online algorithm (\Cref{sec:online algorithm}) to reweight the worst performing class during training and select points that improve its performance. See \Cref{alg:robust active subsampling} for a full description of the REDUCR method.
% We introduce $\beta(x,y,c)$ as a notation shorthand.
% \begin{align}
% \beta (x,y,c) = \max(0, \mathcal{L}[y|x, \theta_t] - \mathcal{L}[y|x, \phi_c])
% \end{align}
%\setlength{\textfloatsep}{5pt}

At each iteration, the top $k$ points are selected (Line 6) according to the weighted sum of  \Cref{eq:robust_rho_loss_restated} for each class $c \in C$, thus efficiently approximating the combinatorial problem from \Cref{eq:online weighted selection rule}. As the class-holdout loss does not depend on the selected points $b_t$ and we sum over the classes, we can remove this term from the weighted sum of the selection scores and only apply it when updating the weights $\mathbf{w}_t$ (in Line 7 and 8). We calculate the \emph{average} class-holdout loss to remove any dependence of the term upon the size of the classes in the holdout dataset. 
We find that clipping the excess loss improves the stability of the algorithm in practice. We test this heuristic empirically in \Cref{sec:Ablation Study} and provide an intuitive explanation for why this is the case in \Cref{app:clipped excess loss ablation}.

When comparing \algname to other online batch selection methods, we observe distinct batch selection patterns. When the dataset is class-imbalanced, the underrepresented classes tend to perform worse because of the lack of training data from those classes. \textsc{RHO-Loss} may struggle to select points from the underrepresented classes as they have less effect on the loss of the holdout dataset. Selection rules that select points with high training loss~\citep{loshchilov2015online,kawaguchi2020ordered,jiang2019accelerating} might select points from the underrepresented classes but have no reference model to determine which of these points are learnable given more data and thus noisy or task-irrelevant points may be selected. In contrast, \algname addresses both of these issues by identifying underrepresented classes and using the class-irreducible loss model to help to determine which points from these classes should be selected.

Even when the dataset is not imbalanced, certain classes might be difficult to learn; for example, due to noise sources in the data collection processes. Via \Cref{eq:weight update},~\algname is able to re-weight the selection scores such that points that are harder to learn from worse-performing classes are selected over points that are easier to learn from classes that are already performing well. This is in contrast to \textsc{RHO-Loss} which will always select points that are easier to learn. We empirically demonstrate this on class balanced datasets in \Cref{sec:experiments}.  

\setlength{\textfloatsep}{5pt}

\begin{algorithm}[t!]
\caption{\algname for robust online batch selection} 
\label{alg:robust active subsampling}
\begin{algorithmic}[1]
    \State \textbf{Input:} 
    data pool $\mathcal{D},$
    holdout data $\mathcal{D}_{ho} = \bigcup_{c \in C} \mathcal{D}_{ho}^{(c)}$, 
    learning rate $\eta \in (0, \infty)$, small batch size $k$, total timesteps $T/k$

    \State Initialize class weights $\mathbf{w}_{1} =\tfrac{1}{C} \mathbf{1}_{C}$

    \State Use $\mathcal{D}_{ho}$ to train $C$ amortised class irreducible loss models as per \Cref{eq:amortised} to obtain $\phi_c$
    %%TRAINING CLASS IRREDUCIBLE LOSS MODEL HERE:
        
    %%FOR LOOP STARTS HERE:
    \For{$t \; \in \; [T/k] $}
    
    \State Receive batch $B_t \subset \mathcal{D}$
    \State $b_t = \underset{b \subset B_t: |b| = k}{\mathrm{argmax}} \sum_{(x,y) \in b} \sum_{c \in C} w_{t,c} \cdot \max(0, \mathcal{L}[y|x, \theta_t] - \mathcal{L}[y|x, \phi_c])$ \Comment{Select points with top k selection scores}
        
    \State Compute the objective value for every class $c \in C$: 
    
    $\alpha_{c} = \sum_{(x,y) \in b_t} \max(0, \mathcal{L}[y|x, \theta_t] - \mathcal{L}[y|x, \phi_c]) - \mathcal{L}[\mathbf{y}_{ho}^{(c)}|\mathbf{x}_{ho}^{(c)}, \theta_{t}] $

    \State Update class weights for every class $c \in C$: $w_{t+1,c} = w_{t,c}\frac{\exp(-\eta \alpha_{c})}{ \sum_{j \in C}w_{t,j}\exp(-\eta \alpha_{j})}$
    
    \State $\theta_{t+1} \leftarrow SGD(\theta_{t}, b_t)$
    \EndFor
\end{algorithmic}
\end{algorithm}

%\vspace{-3pt}
\section{Experiments}
%\vspace{-3pt}
\label{sec:experiments}
In this section, we present empirical results to showcase the performance of \algname on large-scale vision and text classification tasks.
%\subsection{Datasets and Models}
%\vspace{-1.5em}

\textbf{Datasets.} We train and test \algname on image and text datasets. We use CIFAR10 \citep{krizhevsky2009learning}, CINIC10 \citep{darlow2018cinic}, Clothing1M \citep{xiao2015learning}, the Multi-Genre Natural Language Interface (MNLI), and the Quora Question Pairs (QQP) datasets from the GLUE NLP benchmark \citep{wang2018glue}. Each dataset is split into a labelled training, validation and test dataset (for details see \Cref{app:experiment details}), the validation dataset is used to train the class-irreducible loss models and evaluate the class-holdout loss during training. The Clothing1M dataset uses 100k additional points from the training dataset along with the validation dataset to train the irreducible loss model(s) (as per \citep{mindermann2022prioritized}). We simulate the streaming setting by randomly sampling batch $B_t$ from dataset $\mathcal D$ at each timestep.\looseness=-1

% \vspace{-1.5em}
\textbf{Models.} For the experiments on image datasets (CIFAR10, CINIC10 and Clothing1M) all models use a ResNet-18 model architecture \citep{he2016deep}. For the Clothing1M dataset we use a ResNet-18 model pretrained on the imagenet dataset \citep{deng2009imagenet}. The networks are optimised with AdamW \citep{loshchilov2017decoupled} and the default Pytorch hyperparameters are used for all methods except CINIC10 for which the weight decay is set to a value of 0.1. For the NLP dataset we use the \textit{bert-base-uncased} \citep{devlin2018bert} model from HuggingFace \citep{wolf2020transformers} and set the optimizer learning rate to $1e^{-6}$. 
%\vspace{-1.5em}

\textbf{Baselines.} We benchmark our method against the state-of-the-art \textsc{RHO-Loss} \citep{mindermann2022prioritized} and \citet{loshchilov2015online}, an online batch selection method that uses the training loss to select points. We refer to the latter baseline as \textsc{Train Loss}. We also compare against \textsc{Uniform} where points are chosen at random from the large batch at each training step.\footnote{We use training step and timestep interchangeably.} All experiments are run multiple times and the mean and standard deviation across runs calculated. Unless stated otherwise $10\%$ of batch $B_t$ is selected as the small batch $b_t$, and we set $\eta=1e-4$. $\gamma=9$ is used when training each of the amortised class-irreducible loss models on the vision datasets and $\gamma=4$ for the NLP datasets. We study the impact of $\gamma$ and $\eta$ on \algname further in \Cref{sec:hyperparameter tuning}. For full details of the experimental setup see \Cref{app:experiment details}.\footnote{Code available at \href{https://github.com/williambankes/REDUCR}{https://github.com/williambankes/REDUCR}}\looseness=-1 

% Added in response to Reviewer 75mt
\textbf{Metrics.} Finally, it is important to note that we analyse the worst-class test accuracy metric which can be interpreted as a lower bound on a model’s performance under all class distribution shifts. This is because the worst possible distribution shift between the training and test set is one where the entire test set consists of only points from the worst performing class.\looseness=-1

\begin{figure}[t!]
     \centering
     % \begin{subfigure}[t]{0.35\textwidth}
     %     \centering
     %     \includegraphics[width=\textwidth]{Clothing1M_average.pdf}
     %     \caption{Clothing1M average test accuracy}
     %     \label{fig:clothing1m average test accuracy}
     % \end{subfigure}
     \begin{subfigure}[t]{0.33\textwidth}
         \centering
         \includegraphics[width=\textwidth]{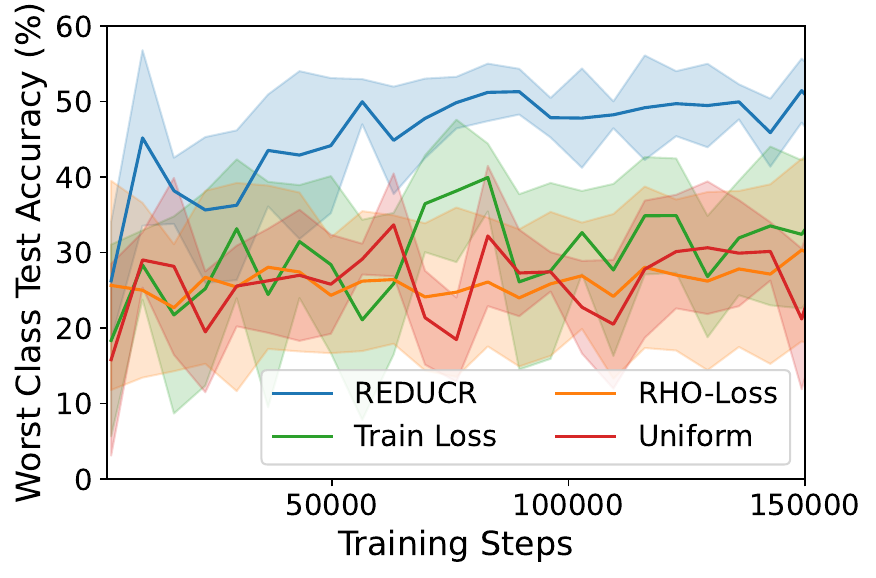}
         \caption{Clothing1M\looseness=-1}
         \label{fig:clothing1m worst class test accuracy}
     \end{subfigure}
     \begin{subfigure}[t]{0.33\textwidth}
         \centering
         \includegraphics[width=\textwidth]{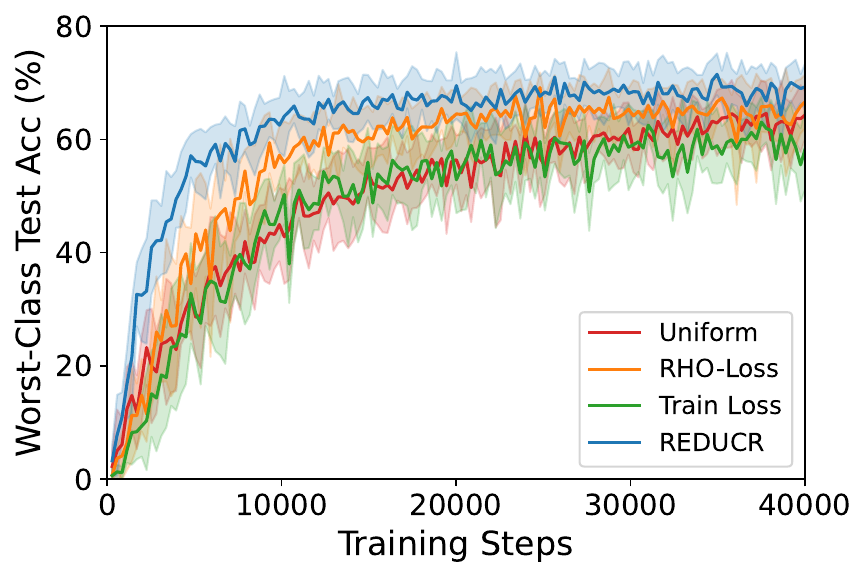}
         \caption{CINIC10 \looseness=-1}
         \label{fig:cinic10 worst class test accuracy}
     \end{subfigure}
     \begin{subfigure}[t]{0.32\textwidth}
         \centering
         \includegraphics[width=\textwidth]{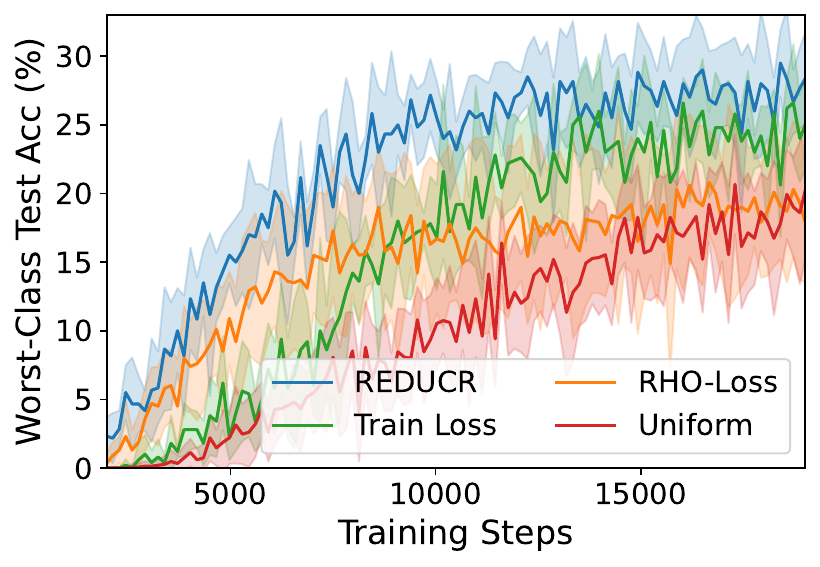}
         \caption{CIFAR100\looseness=-1}
         \label{fig:cifar100 worst class test accuracy}
     \end{subfigure}
        \caption{\algname improves the worst-class test accuracy and data efficiency when compared with the \textsc{RHO-Loss}, \textsc{Train Loss} and \textsc{Uniform} baselines on the \subref{fig:clothing1m worst class test accuracy}) Clothing1M dataset,~\subref{fig:cinic10 worst class test accuracy}) the CINIC10 dataset, and~\subref{fig:cifar100 worst class test accuracy}) the CIFAR100 dataset.\looseness=-1}
        \label{fig:clothing1m results}
\end{figure}

\begin{table*}[t!]
\caption{\algname outperforms RHO-LOSS (the best overall baseline) in terms of the worst-class test accuracy on Clothing1M, CINIC10 and CIFAR10 by at least ~5-26\%. Across all baselines, \algname gains about 15\% more accuracy on the noisy and imbalanced Clothing1M dataset as shown in \Cref{fig:clothing1m worst class}.$\phantom{1}^{*}$CIFAR100 results from training step 10k where \algname converges, after 10k further training steps \textsc{Train Loss} achieves a similar performance.}
%\vspace{-0.7em}
\begin{center}
\begin{tabular}{lllll}
\hline
\multirow{2}{*}{\bf Dataset} &\multicolumn{4}{c}{\bf Worst-Class Test Accuracy (\%) $\pm 1$ std} 
\\

%\cline{2-7}
& \textbf{\textsc{Uniform}} & \textbf{\textsc{Train Loss}} &\textbf{\textsc{RHO-Loss}} & \textbf{\algname}
\\ \hline \\

%Just creating the table as a place holder
CIFAR$10$ (10 runs) & 75.01 $\pm$ 1.37 & 76.1 $\pm$ 2.31 & 
 78.80 $\pm$ 2.09 & \textbf{83.29} $\pm$ \textbf{0.84} \\
CINIC$10$ (10 runs) & 64.70 $\pm$ 2.45 & 64.83 $\pm$ 4.75 &
 69.39 $\pm$ 3.56 & \textbf{75.30} $\pm$ \textbf{0.85} \\
CIFAR$100^*$ (5 runs) & 10.59 $\pm$ 3.63 & 17.59 $\pm$ 5.17 & 16.0 $\pm$ 6.93 & \textbf{26.00} $\pm$ \textbf{2.65} \\
Clothing1M (5 runs) & 39.23 $\pm$ 5.41 & 40.37 $\pm$ 3.58 & 27.77 $\pm$ 10.16 & \textbf{53.91} $\pm$ \textbf{2.42}\\
MNLI (5 runs) & 74.70 $\pm$ 1.26 & 74.56 $\pm$ 1.44 & 76.74 $\pm$ 0.93 & \textbf{79.45} $\pm$
\textbf{0.39} \\
QQP (5 runs) & 73.21 $\pm$ 2.04 & 79.96 $\pm$ 2.34 & 78.21 $\pm$ 1.95 &	\textbf{86.61} $\pm$ \textbf{0.49} \\
\end{tabular}
\end{center}
     \vspace{-5pt}
\label{tab:worst class results average checkpoint}
\end{table*}
\begin{table*}[t!]
\caption{Together with \Cref{tab:worst class results average checkpoint}, these results demonstrate that \algname improves the worst-class test accuracy while maintaining strong average test accuracy despite \algname not explicitly optimizing the average test accuracy. \algname outperforms the baseline methods on the CIFAR$100$ and Clothing1M datasets.}
%\vspace{-0.7em}
\begin{center}
\begin{tabular}{lllll}
\hline
\multirow{2}{*}{\bf Dataset} &\multicolumn{3}{c}{\bf Average Test Accuracy (\%) $\pm 1$ std} 
\\

%\cline{2-7}
& \textbf{\textsc{Uniform}} & \textbf{\textsc{Train Loss}} &\textbf{\textsc{RHO-Loss}} & \textbf{\algname}
\\ \hline \\

%Just creating the table as a place holder
CIFAR$10$ (10 runs) & 85.09 $\pm$ 0.52 & 88.86 $\pm$ 0.22 & \textbf{90.00} $\pm$ \textbf{0.33} &  \textbf{90.02} $\pm$ \textbf{0.44} \\
CINIC$10$ (10 runs) & 79.51 $\pm$ 0.30 & 79.25 $\pm$ 0.33 & \textbf{82.09} $\pm$ \textbf{0.30} & \textbf{81.68} $\pm$ \textbf{0.47} \\
CIFAR$100$ (5 runs) & 57.94 $\pm$ 0.69 & 59.77 $\pm$ 0.71 & 60.95 $\pm$ 0.64 & 
\textbf{62.21} $\pm$ \textbf{0.62} \\
Clothing1M (5 runs) & 69.60 $\pm$ 0.85 & 69.63 $\pm$ 0.30 & 71.07 $\pm$ 0.46 & \textbf{72.69} $\pm$ \textbf{0.42}\\
MNLI (5 runs) & 79.19 $\pm$ 0.53 & 76.85 $\pm$ 0.14 & \textbf{80.89} $\pm$ \textbf{0.31} & \textbf{80.28} $\pm$ \textbf{0.33} \\
QQP (5 runs) & 85.05 $\pm$ 0.43 & \textbf{86.30} $\pm$ \textbf{0.41} & \textbf{86.88} $\pm$ \textbf{0.31} & \textbf{86.99} $\pm$ \textbf{0.49} \\
\end{tabular}
\end{center}
     %\vspace{-1em}
%\caption{\algname matches or outperforms the average test accuracy of the best competing baseline across all datasets. Note that optimizing the average test accuracy is not the objective of \algname. These results, together with \Cref{tab:worst class results average checkpoint}, demonstrate the significant advantage of \algname to improve the worst-class accuracy while maintaining the strong average-case performance. }

\label{tab:average results average checkpoint}
\end{table*}

\subsection{Key results}
%\vspace{-10pt}

The worst-class and average test accuracy for the datasets and model are shown in \Cref{tab:worst class results average checkpoint} and \Cref{tab:average results average checkpoint}, respectively. 
Across all datasets, \algname outperforms the baselines in terms of the worst-class accuracy and matches or even outperforms the average test accuracy of \textsc{RHO-Loss} within one standard deviation. This is surprising because the primary goal of \algname is not to optimize the overall average (over classes) performance.
\algname performs particularly strongly on the Clothing1M dataset, \Cref{tab:worst class results average checkpoint} shows \algname improves the worst-class test accuracy by around $15\%$ when compared to \textsc{Train Loss}, the next best-performing baseline, and by around 26\% when compared to \textsc{RHO-Loss}, the overall best-performing baseline across datasets. \Cref{fig:clothing1m worst class test accuracy} shows that \algname also achieves this performance in a more data efficient manner than the comparable baselines, achieving a mean worst-class test accuracy of $40\%$ within the first 10k training steps. We also observe improved efficiency on the CINIC10 dataset, as shown in \Cref{fig:cinic10 worst class test accuracy}, and the MNLI and QQP datasets as detailed in \Cref{fig:qqp and mnli results}.\looseness=-1

\begin{wrapfigure}{R}{0.45\textwidth}
    %\vspace{-1em}
         \centering
         \includegraphics[width=0.4\textwidth]{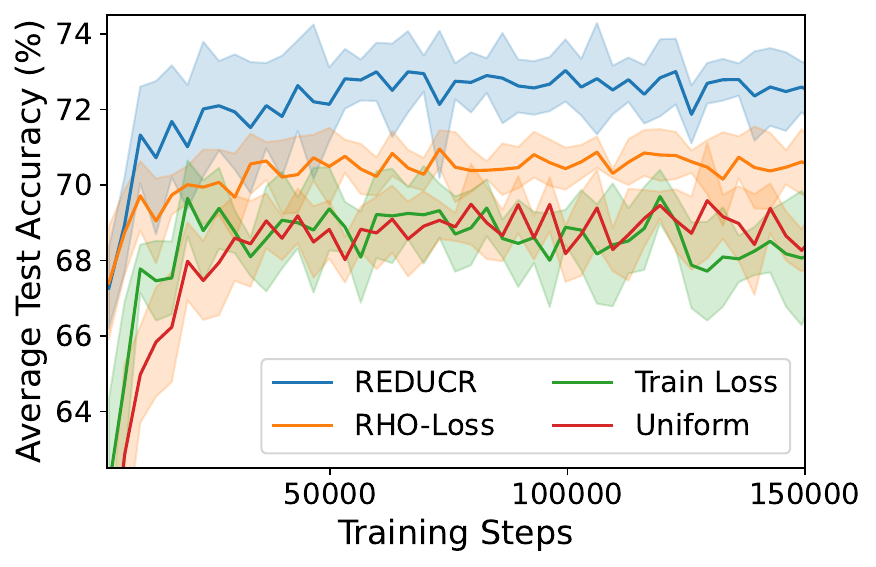}
         \caption{REDUCR improves the average test accuracy on the Clothing1M dataset.}
         \label{fig:clothing1m average test accuracy}
    %\vspace{-3em}
\end{wrapfigure}

The Clothing1M dataset also sees a distribution shift between the training and test dataset. In the test dataset, the worst performing class is much more prevalent than in the training dataset and as such improvements to its performance impact the average test accuracy significantly. \Cref{fig:clothing1m average test accuracy} shows the impact of this distribution shift as the improved performance of the model on the worst-class results in an improved average test accuracy to the state-of-the-art \textsc{RHO-Loss} baseline.
%\vspace{-3pt}
\subsection{Ablation Studies}
%\vspace{-3pt}
\label{sec:Ablation Study}

\begin{figure*}[t!]
     \centering
     \begin{subfigure}[t]{0.32\textwidth}
         \centering
         \includegraphics[width=\textwidth]{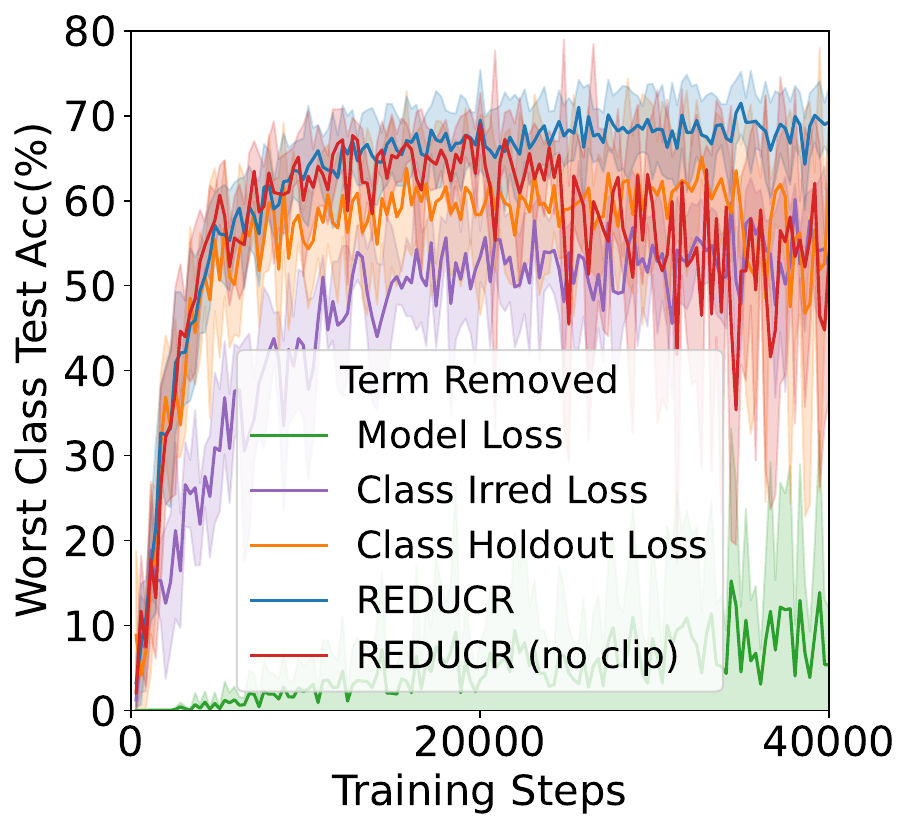}
         \caption{}
         \label{fig:cinic10_ablation_accuracy}
     \end{subfigure}
     \hfill
     \begin{subfigure}[t]{0.32\textwidth}
         \centering
         \includegraphics[width=\textwidth]{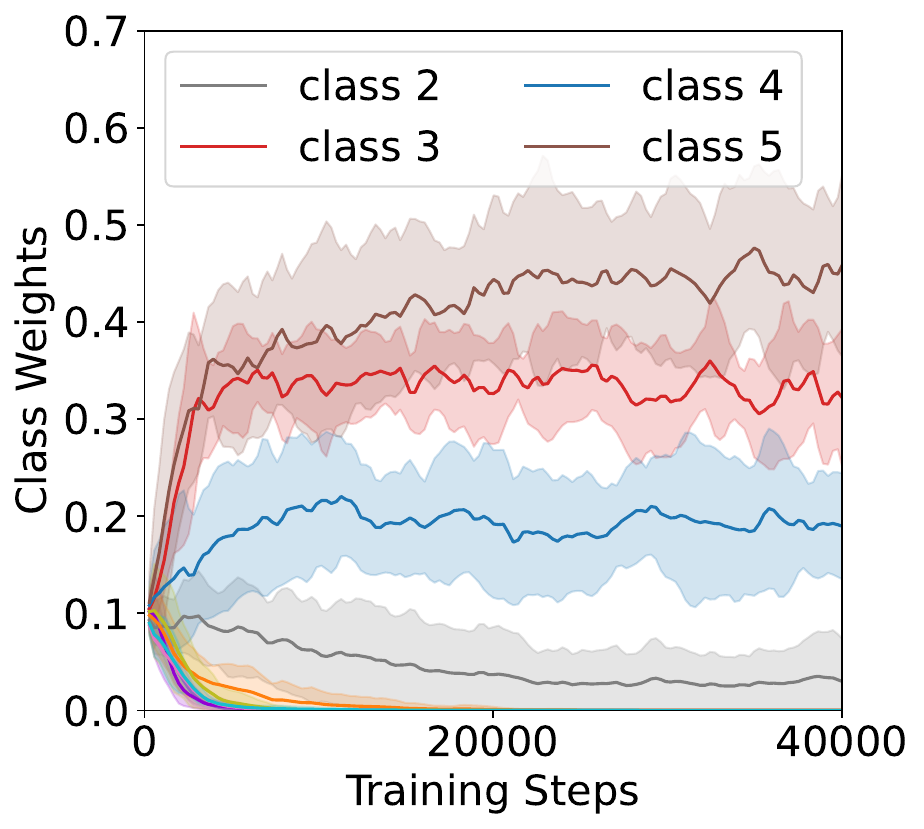}
         \caption{}
         \label{fig:cinic10 weights}
     \end{subfigure}
     \begin{subfigure}[t]{0.32\textwidth}
         \centering
         \includegraphics[width=\textwidth]{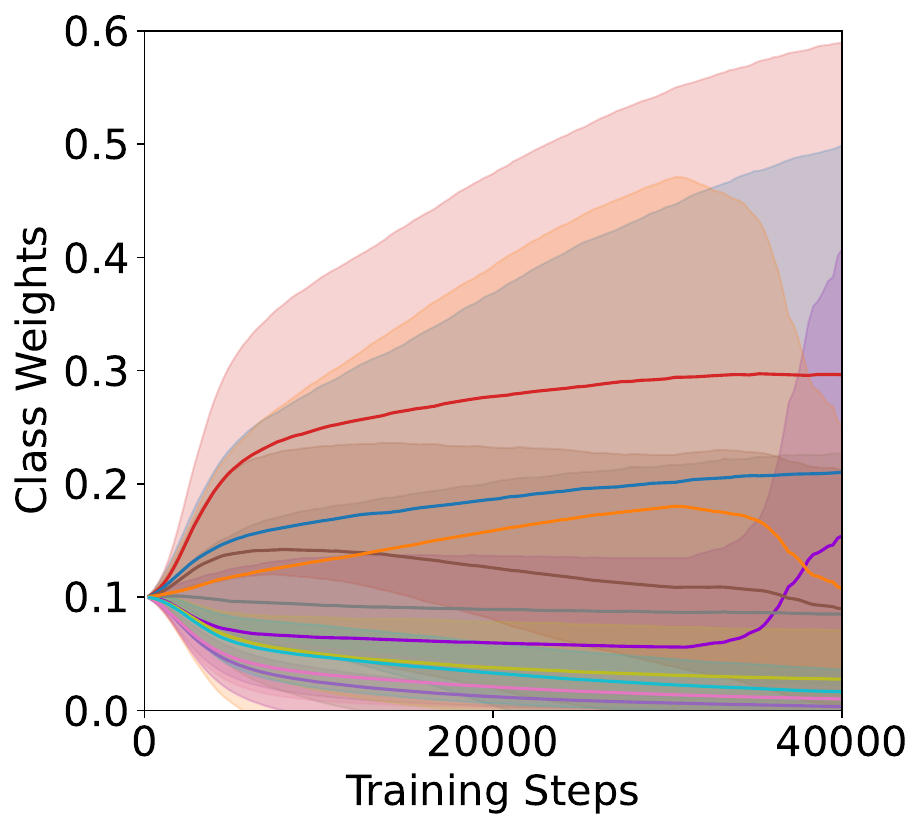}
         \caption{}
         \label{fig:cinic10 weights ablation}
     \end{subfigure}
        \caption{\subref{fig:cinic10_ablation_accuracy}) The worst-class test accuracy \emph{decreases} when the model loss, class irreducible loss, and class-holdout loss terms are removed from REDUCR on CINIC10. Comparing REDUCR with clipping for excess losses (\Cref{alg:robust active subsampling}) and REDUCR (no clip) which removes the clipping, we observe that REDUCR achieves more stable performance. We show the class weights $\mathbf{w}$ at each training step for \subref{fig:cinic10 weights}) \algname and \subref{fig:cinic10 weights ablation}) \algname with the class-holdout loss term ablated. The ablation model fails to consistently prioritise the underperforming classes across multiple runs. \looseness=-1}
        \label{fig:cinic10 ablation and weight results}
        %\vspace{-10pt}
\end{figure*}

To further motivate the selection rule in \Cref{eq:robust_rho_loss_restated}, we conduct a series of ablation studies to show that all the terms are necessary for robust online batch selection. \Cref{fig:cinic10_ablation_accuracy} shows the performance of \algname on the CINIC10 dataset when the model loss, amortised class-irreducible loss and class-holdout loss terms of the algorithm were individually excluded from the selection rule. All three terms in \Cref{eq:robust_rho_loss_restated} are required to achieve a strong worst-class test accuracy. 

%% Add this into the text somewhere here:
%“In Figure 4a) we note that 
Removing the Model Loss results in the worst performance in the set of ablation studies. This is because the Model Loss provides REDUCR with information about which points are currently not classified correctly by the model. By removing this term REDUCR only selects points which do well under the Class Irreducible Loss model and does not prioritise points the model has not yet learnt. Selecting points not yet learnt by the model is an important quality in online batch selection approaches and the main premise of the Train Loss baseline algorithm.
Likewise by removing the Class Irreducible Loss Model term we remove the ability of the model to infer if a point can be learnt or not. In \cite{mindermann2022prioritized}, the authors note that these pretrained models enable the algorithm to pick points that are learnable and do not have label noise.

The removal of the class-holdout loss term affects the ability of $\algnamens$ to prioritise the weights of the model correctly. In \Cref{fig:cinic10 ablation and weight results} we compare the class weights of~\algname and an ablation model without the class-holdout loss term. The standard model clearly prioritises classes 3, 4 and 5 during training across all 5 runs, whilst the ablation model does not consistently weight the same classes across multiple runs. We also conducted an ablation study on the clipping of the excess loss to motivate its inclusion in the algorithm, this is also shown in \Cref{fig:cinic10_ablation_accuracy}, we note that this stabilises the model performance towards the end of training and investigate further in \Cref{app:clipped excess loss ablation}.   
%\vspace{-10pt}
\subsection{Scaling up the number of classes}
%\vspace{-3pt}
\label{sec: scaling up the number of classes}
%The computational expense of \algname scales with the size of the number of classes $|C|$ as we must train a class-irreducible loss model $\forall c \in C$. One solution is to group the classes into superclasses and solve the robust data downsampling problem over these superclasses. We test this on the CIFAR100 dataset using the provided groupings. The results for this can be seen in \cref{fig:cifar100 worst class test accuracy}, \algname outperforms the baseline models in terms of the worst-\textbf{class} test accuracy.
\algname can handle problems involving a large number of classes without needing to train a separate class-irreducible loss model for each class. One idea is to group the classes into superclasses, where $c \in \mathcal{G}_{i}, \mathcal{G}_{i} \in \{\mathcal{G}_i\}_{i=1}^{|G|}$, $\mathcal{G}_{i} \cap \mathcal{G}_{j} = \emptyset$ for $i \neq j$ and $|G| < |C|$, and solve the robust data downsampling problem over these superclasses. We test the proposed variant of \algname on the CIFAR100 dataset using the provided groupings with 20 superclasses in total~\citep{krizhevsky2009learning}. \cref{fig:cifar100 worst class test accuracy} shows that \algname outperforms the baselines in terms of the worst-\textbf{class} test accuracy, even though the robust objectives are over the superclasses. It achieves this performance in \textbf{half} the number of training steps as shown in \Cref{tab:worst class results average checkpoint}. \looseness=-1

%\vspace{-3pt}
\subsection{Imbalanced Datasets}
%\vspace{-3pt}

\begin{figure*}
     \centering
     %\vspace{-9.01em}
     \hspace{2em}
     \begin{subfigure}[t]{0.4\textwidth}
         \centering
         \includegraphics[width=\textwidth]{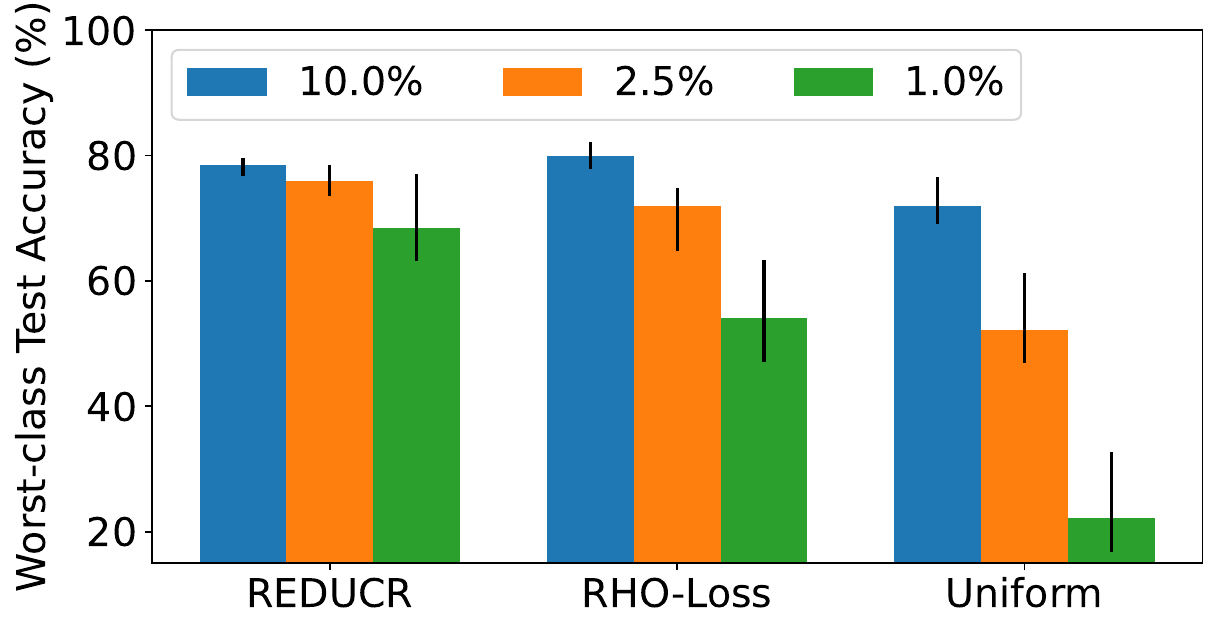}
         \caption{Under-sampling on class 3}
         \label{fig:cifar10 class 3 under-sampling}
     \end{subfigure}
    \hfill
     \begin{subfigure}[t]{0.4\textwidth}
         \centering
         \includegraphics[width=\textwidth]{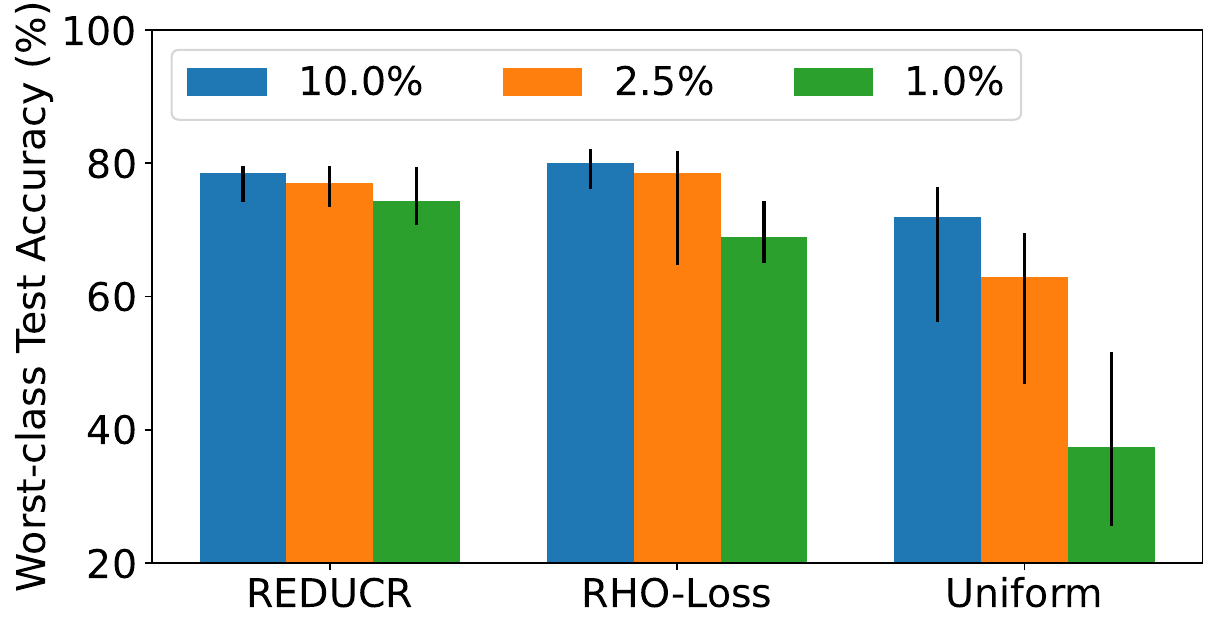}
         \caption{Under-sampling on class 5}
         \label{fig:cifar10 class 5 under-sampling}
     \end{subfigure}
    \hspace{2em}
    \caption {\algname significantly reduces the deterioration in the worst-class test accuracy on CIFAR10 for ~\subref{fig:cifar10 class 3 under-sampling}) class 3 and ~\subref{fig:cifar10 class 5 under-sampling}) class 5  when compared to the \textsc{Uniform} and \textsc{RHO-Loss} baselines as the percent imbalance is decreased from 10.0\% (balanced) to 1.0\%. Each experiment was repeated 10 times, the median value was plotted and the error bars denote the best and worst run across 10 runs.\looseness=-1}
        \label{fig:cifar10 percent imbalance results}
\end{figure*}

We investigate the performance of models trained using \algname on imbalanced datasets. We artificially imbalance the CIFAR10 training and validation datasets such that a datapoint of the imbalanced class is sampled with probability $p \in (0, 1/C]$ (referred to as the percent imbalance) and datapoints from the remaining classes are sampled with probability $(1 - p)/(C - 1)$ during model training. We conduct experiments with 0.01, 0.025 and 0.1 (which is equivalent to the balanced dataset) percent imbalance on classes 3 and 5. The results are shown in \Cref{fig:cifar10 percent imbalance results}. 

We find the performance of models trained using \algname deteriorates less than those trained with the \textsc{RHO-Loss} or \textsc{Uniform} baselines as the percent imbalance of a particular class decreases (see \Cref{fig:cifar10 percent imbalance results}). For example, when class 3 is imbalanced, in the most imbalanced case (1.0\%) the median performance of \algname outperforms that of \textsc{RHO-Loss} run by 14\%. This demonstrates the effectiveness of \algname in prioritising the selection of data points from underrepresented classes.\looseness=-1
%\vspace{-6pt}
\section{Conclusions, Broader Impact and Limitations}
%\vspace{-6pt}
% conclusion paragraph
In summary, we identified the problem of class-robust data downsampling and proposed a new method, \algname, to solve this problem using class priority reweighting. Our experimental results indicate that \algname significantly enhances data efficiency during training, achieving superior test accuracy for the worst-performing class and frequently surpassing state-of-the-art methods in terms of average test accuracy. \algname excels in settings where the available data are class-imbalanced by prioritising the selection of points from underrepresented classes.

\textbf{Limitations.} %Compared with non-robust data selection, 
The computational efficiency of \algname scales linearly with the number of classes. We propose one solution to this in \Cref{sec: scaling up the number of classes}, where we show that using groups of classes can still result in improved worst-class performance. Another solution is to use smaller model architectures for the class-irreducible loss model. \citep{mindermann2022prioritized} provide extensive evidence that small reference models can improve computational efficiency whilst still providing a useful signal for data selection, we leave investigation of these methods as a future research direction.

\textbf{Broader Impact.} Improving data efficiency is an important and practical problem as more machine learning models are being trained and deployed for real-world applications. Moreover it is critical to ensure the robustness of models for reliable and trustworthy machine learning. Our work proposes a new method with the goal of improving the robustness of models whilst significantly reducing the data required to achieve state of the art performance.\looseness=-1

\section*{Acknowledgments}
This work was supported by the EPSRC New Investigator Award EP/X03917X/1; the Engineering and Physical Sciences Research Council EP/S021566/1; Google Research Scholar award and the Google Cloud Platform Credit Award. We thank Zoubin Ghahramani, Jasper Snoek, Fei Sha, Théo Galy-Fajou, Dustin Tran, Sharat Chikkerur, Xuezhi Wang and Mike Dusenberry for helpful discussions on early versions of this work.

% \vspace{-2pt}
% \section*{Broader Impact}
% \vspace{-2pt}
% Improving the data efficiency is a very important and practical problem as more and more machine learning models are being trained and deployed for real-world applications. Moreover, it is critical to ensure the robustness of models for reliable and trustworthy uses of machine learning. Our work proposes a new method with the goal to improve the robustness of models while significantly reduce the carbon footprint of model training. 

%\vspace{-2pt}
%\section*{Reproducibility}
%\vspace{-2pt}
% All the experiments mentioned in the Paper and Appendix were implemented using the code provided in \url{https://anonymous.4open.science/r/REDUCR-24D3} which includes the necessary code for processing the raw datasets, which are freely available online.\looseness=-1

%\bibliography{Styles/references}
\bibliography{references}

\appendix
\onecolumn

\section{Appendix}

\subsection{Online Batch Selection Pseudo-Code}
\label{sec:online batch selection pseudo code}
For the sake of convenience, we provide the pseudocode of the online batch selection protocol described in \Cref{sec:online_batch_selection_subsection}.
\begin{algorithm}[h!]
\caption{Online batch selection}\label{alg:online batch selection}
\begin{algorithmic}[1]
    \State \textbf{Input:} 
    data pool $\mathcal{D},$
    number of training steps $T$,
    stochastic gradient descent algorithm \textsc{SGD},
    a loss function $\mathcal{L}$

    %%FOR LOOP STARTS HERE:
    \For{$t = 1 \; \text{to} \; T$}
    \State Sample batch $B_t$ randomly from $\mathcal{D}$

    \State $b_t = \text{SelectBatch}(B_t, \theta_t)$
    \State L = $ \sum_{(x_i, y_i) \in b_t} \mathcal{L}[y_i| x_i, \theta_t]$

    \State $\theta_{t+1} = \textsc{SGD}(L, \theta_t)$

    \EndFor
\end{algorithmic}
\end{algorithm}
\subsection{Class Irreducible Loss Model Training Pseudo-Code}
\label{app:class irred model training}
Here we detail the pseudo-code for training the class irreducible loss model described in \Cref{sec:class irred loss model}
\begin{algorithm}[h!]
\caption{Class Reference Model Training}\label{alg:class irred model training}
\begin{algorithmic}[1]

    \State \textbf{Input:} 
    holdout dataset $\mathcal{D}_{ho},$
    number of training steps $T$,
    stochastic gradient descent algorithm \textsc{SGD},
    a loss function $\mathcal{L}$, a specific class $c$
    
    %%FOR LOOP STARTS HERE:    
    \For{$t = 1 \; \text{to} \; T$}
    \State $B_{ho} \sim \text{Uniform}(\mathcal{D}_{ho})$

    \State $L = \sum_{(x_i,y_i) \in B_{ho}} (1 + 
 \gamma \mathbb{I}[c = y]) \mathcal{L}[y_i |x_i, \phi_t]$

    \State $\phi_{t+1} = \textsc{SGD}(L, \phi_t)$

\EndFor

\State \textbf{Return} Class-irreducible loss model parameters $\phi_{T}$

\end{algorithmic}
\end{algorithm}

\subsection{The Amortised Class Irreducible Loss Model Approximation}
\label{app:class irred model approx}

The amortised class irreducible loss model is an important component in \algname as shown by our ablation study in \Cref{fig:cinic10_ablation_accuracy}. For each class $c \in [C]$, we approximate the second term of \Cref{eq:Bayes_rule}, $\log p(y|x, \mathcal{D}_{t}, \mathcal{D}_{ho}^{(c)})$ via the model trained using \Cref{alg:class irred model training}. This approximation has two steps: firstly we remove dependence of the class irreducible model loss on the training dataset at time $t$. A similar approximation is heavily explored by \citet{mindermann2022prioritized} in Section 3 and Section 4 of their paper; in Appendix D of their work they show that this approximation is important for the stable training of \textsc{RHO-Loss}. The approximation also aligns \textsc{RHO-Loss} and \algname with other methods in the literature such as \citet{xie2023doremi, oren2019distributionally} which similarly use a reference model that does not vary during training. 

Secondly we up-weight data points in the loss function when their label $y \in [C]$ matches that of the specific class $c$. Unlike \textsc{RHO-Loss} we cannot approximate the class irreducible loss as $\log p(y|x, \mathcal{D}_{t}, \mathcal{D}_{ho}^{(c)}) \approx \log p (y|x, \mathcal{D}_{ho}^{(c)})$ as this is a trivial model only trained on points with labels from a single class and thus does not provide a suitable signal to guide point selection. We interpret the original class irreducible loss $\log p(y|x, \mathcal{D}_{t}, \mathcal{D}_{ho}^{(c)})$, as an expert model for class $c$ as this model trains on extra points only sampled from that class, $\mathcal{D}_{ho}^{(c)}$. In our approximation we train on the holdout dataset which does not have excess examples of points from class $c$. We justify our up-weighting of points as a form of importance weighting \citep{shimodaira2000improving}, where by up-weighting points with labels in a specific class we are calculating a Monte Carlo approximation of the loss under a distribution in which points from class $c$ are more prevalent.

\subsection{The Effect of the Class-Holdout Loss on the Selection of Points}
\label{sec:class holdout loss appendix}

The class-holdout loss term only affects the selection of points at each iteration $t$ through the selection of the weights $\mathbf{w}_{t}$. As it does not depend upon the candidate point $(x,y) \in B_{t}$ and the weights sum to one we can remove it from line 6 of \Cref{alg:robust active subsampling} and only include it in line 7 when we update the class weights. Similarly as the model loss does not depend upon the class $c$ we can write the selection score as 
\begin{equation}
    \sum_{c \in C} w_{t,c} \log p(\yhoc|\xhoc, \mathcal{D}_{t} \cup (\{x, y\})) = \\\targetl - \sum_{c \in C}w_{t,c} (\irredlc) - \sum_{c \in C}w_{t,c} (\robusthl).
\end{equation}  

\subsection{Experiment Details}
\label{app:experiment details}

\begin{figure}[t!]
     \centering
     \begin{subfigure}[t]{0.49\textwidth}
         \centering
         \includegraphics[width=\textwidth]{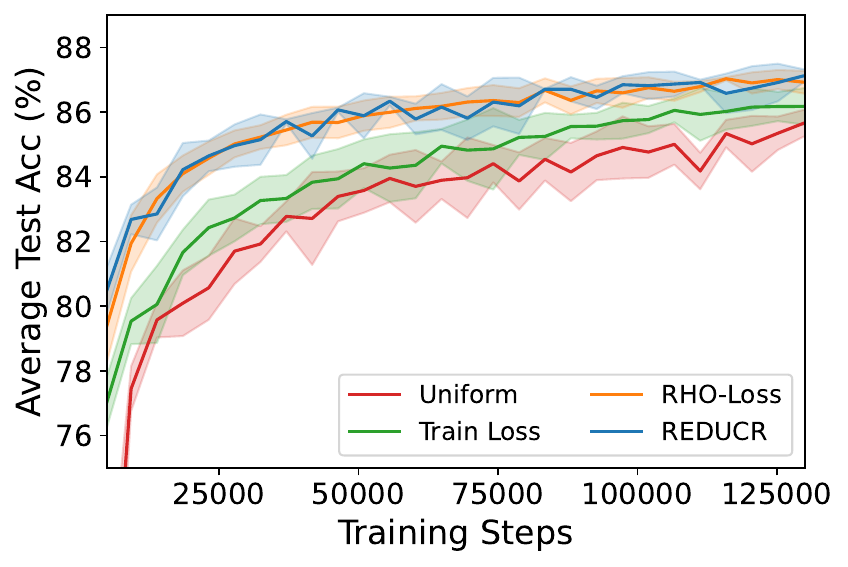}
         \caption{QQP average test accuracy}
         \label{fig:qqp average test accuracy}
     \end{subfigure}
     \hfill
     \begin{subfigure}[t]{0.49\textwidth}
         \centering
         \includegraphics[width=\textwidth]{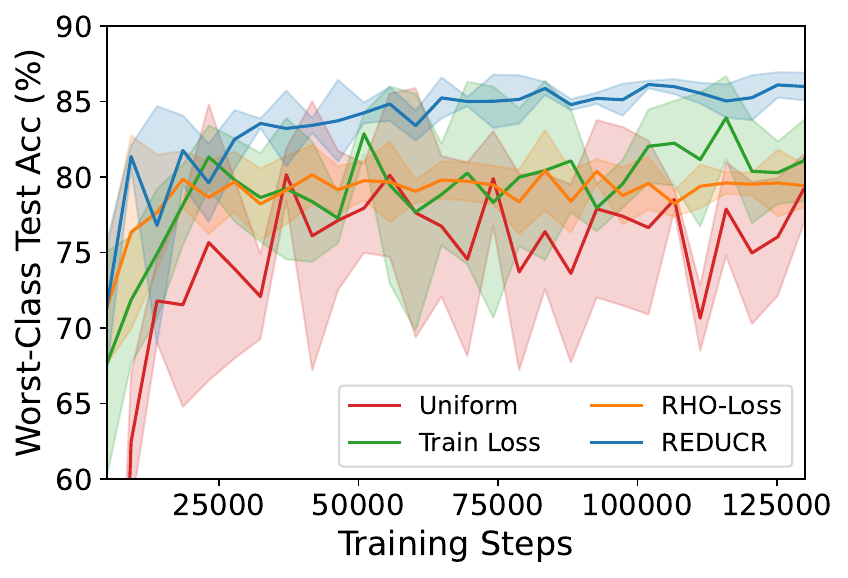}
         \caption{QQP worst-class test accuracy}
         \label{fig:qqp worst class test accuracy}
     \end{subfigure}
     \begin{subfigure}[t]{0.49\textwidth}
         \centering
         \includegraphics[width=\textwidth]{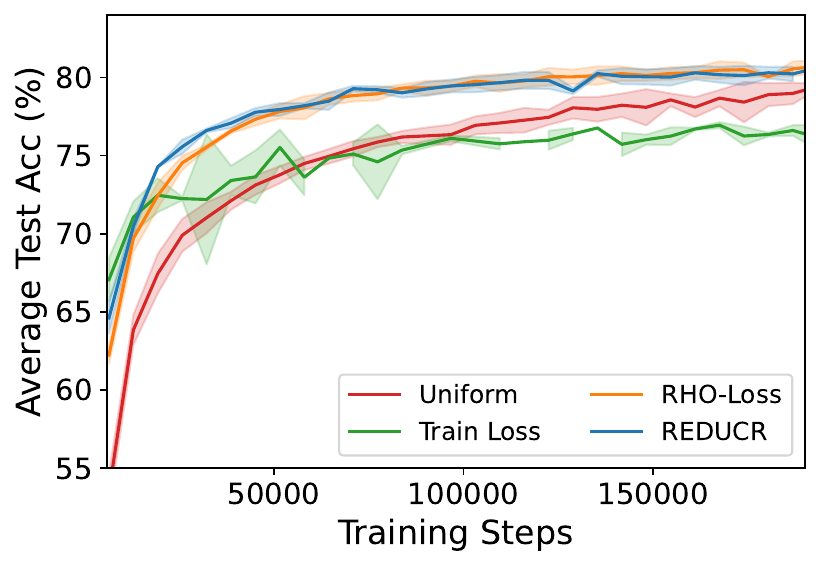}
         \caption{MNLI average test accuracy}
         \label{fig:mnli average test accuracy}
     \end{subfigure}
     \hfill
     \begin{subfigure}[t]{0.49\textwidth}
         \centering
         \includegraphics[width=\textwidth]{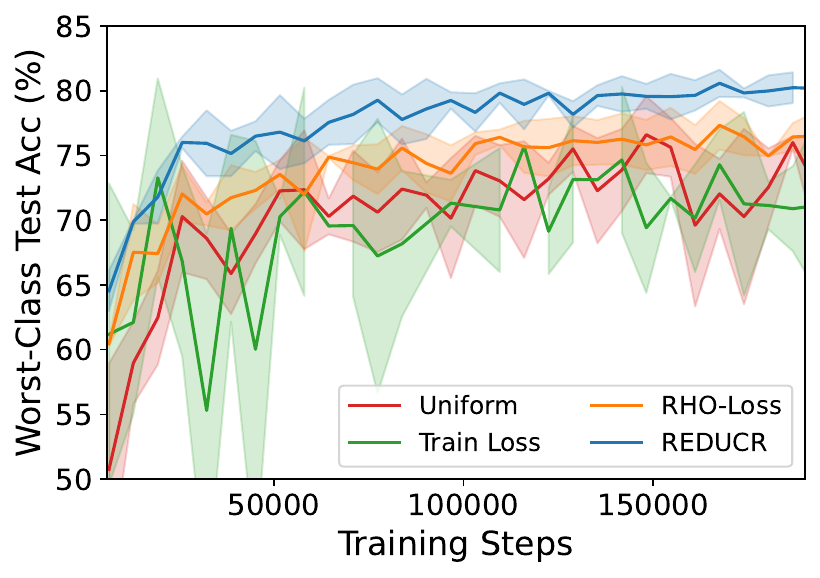}
         \caption{MNLI worst-class test accuracy}
         \label{fig:mnli worst class test accuracy}
     \end{subfigure}
        \caption{\algname improves the worst-class test accuracy on the MNLI and QQP text datasets whilst maintaining strong average test accuracy performance when compared with the \textsc{Train Loss}, \textsc{RHO-Loss} and \textsc{Uniform} baselines. On both datasets \algname matches the next best performing baseline's mean result across runs approximately 100k training steps earlier.}
        \label{fig:qqp and mnli results}
\end{figure}

We provide the full code base anonymised for review purposes as part of the supplementary material.\looseness=-1 

\textbf{CIFAR10} used half the training dataset (25k points) as a holdout validation dataset for training the amortised class-irreducible loss models and calculating the class-holdout loss during the robust online batch selection. We used the remaining 25k points as a training dataset and the provided test dataset (10k) for testing.

\textbf{CINIC10} used the provided validation dataset for both the class holdout loss and amortised class irreducible loss models.

\textbf{CIFAR100} used the provided validation dataset for both the class holdout loss and amortised class irreducible loss models.

\begin{wrapfigure}{R}{0.45\textwidth}
    \vspace{-1.5em} 
    \centering  \includegraphics[width=0.4\textwidth]{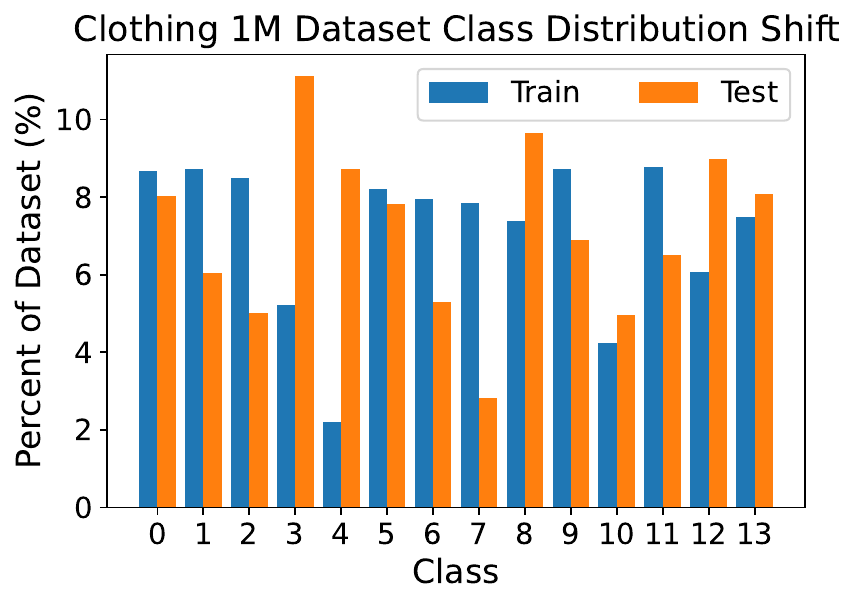}
    %\vspace{-5pt}
     \caption{Clothing1M train-test class distribution shift. The number of points in classes 4 and 7 change dramatically between the train and test sets.\looseness=-1} 
      \label{fig:clothing1m_dist_shift}
      \vspace{-2em}
\end{wrapfigure}

\textbf{Clothing1M.} The dataset consists of 1 million images labelled automatically using the keywords in its surrounding text. The dataset consists of 72k 'clean' images whose labels have been hand checked, 50k, 13k and 9k are respectively sorted into a clean training, validation, and test sub-dataset. To train the amortised class irreducible loss models we use 100k points randomly sampled from the union of the validation, clean and noisy training datasets. We calculate the class-holdout loss term and validation performance during training using the clean validation dataset. \Cref{fig:clothing1m_dist_shift} shows how the class distribution changes between train and test times. 

\textbf{MNLI.} The dataset \citep{williams2017broad} consists of 412k labeled sentence pairs; similarly to \citet{sagawa2019distributionally} we split these sentence pairs into a train (206k), validation (164k), and labelled test (41k) dataset.

\textbf{QQP.} The dataset consists of 431k labeled sentence pairs; we remove points from class 1 to further imbalance the dataset resulting in 22\% of the dataset labelled class 1. We split the remaining points into a train (148k), validation (67k), and labelled test (40k) dataset. We do not adjust the balance of the test dataset.

\textbf{ResNet-18} used for the Clothing1M experiments is the pretrained model available via the Torchvision~\citep{marcel2010torchvision} model library. For the CIFAR10 and CINIC10 experiments we use the adapted ResNet-18 architecture detailed in~\citet{mindermann2022prioritized} Appendix B. 

\textbf{Train Loss} baseline is taken from \citet{loshchilov2015online} where points from the large batch $B_t$ are sampled with probability
\begin{equation}
    p_i \propto \frac{1}{\exp(\log(s)/|B_t|)^{i}}.
\end{equation}
Here $p_i$ is the point with the $i^{th}$ highest training loss in the large batch. We set the selection pressure parameter $s_e = 100$ and do not vary this during training as per the Experiments in Section 6. of \citet{loshchilov2015online}. 

\textbf{Compute Resources and Data Sources.} All models were trained on GCP NVIDIA Tesla T4 GPUs. The Image datasets were sourced from pytorch via the torchvision datasets package \url{https://pytorch.org/vision/stable/datasets.html}, the NLP datasets were sourced from huggingface, \url{https://huggingface.co/datasets/nyu-mll/glue}.  

\textbf{Data Augmentation} was applied to the training dataset during online batch selection and validation dataset during the training of the amortised class-irreducible loss model. We apply a random crop and random flip to the images.

\subsection{Results with worst-class checkpointing}

In \Cref{tab:worst class results worst class chkp} and \Cref{tab:average results worst class chkp} we show the worst-class and average test accuracy respectively, when the \textsc{Uniform}, \textsc{Train Loss} and \textsc{RHO-Loss} baselines use worst-class validation accuracy to checkpoint the model during training. \algname still outperforms or matches the best baseline performance across all datasets. In the cases where \algname matches the performance of the best performing baseline, it does so in a more data efficient manner. \Cref{fig:qqp worst class test accuracy} and \Cref{fig:mnli worst class test accuracy} show the mean and standard deviation worst-class test accuracy across multiple runs on the QQP and MNLI datasets. \algname matches the best mean performance of the best performing baseline almost 100k training steps earlier on both datasets.    

\begin{table}[h!]
\begin{center}
\begin{tabular}{lllll}
\hline
\multirow{2}{*}{\bf Dataset} &\multicolumn{4}{c}{\bf Worst-Class Test Accuracy (\%) $\pm 1$ std} 
\\

%\cline{2-7}
& \textbf{\textsc{Uniform}} & \textbf{\textsc{Train Loss}} &\textbf{\textsc{RHO-Loss}} & \textbf{\algname}
\\ \hline \\

%Just creating the table as a place holder
CIFAR10 (10 runs) & 75.01 $\pm$ 1.37 & 79.32 $\pm$ 1.35 & 81.23 $\pm$ 1.18 & \textbf{83.29} $\pm$ \textbf{0.84} \\
CINIC10 (10 runs) & 70.86 $\pm$ 1.23 & 68.89 $\pm$ 0. 86 &
 \textbf{73.44} $\pm$ \textbf{1.16} & \textbf{75.30} $\pm$ \textbf{0.85} \\
Clothing1M (5 runs) & 39.23 $\pm$ 5.41 & 49.02 $\pm$
  2.32 & 32.19 $\pm$ 9.83 & \textbf{53.91} $\pm$ \textbf{2.42}\\
MNLI (5 runs) & 76.88 $\pm$ 1.21 & 75.75 $\pm$ 0.56 & \textbf{78.04} $\pm$ \textbf{1.73} & \textbf{79.45} $\pm$ \textbf{0.39} \\
QQP (5 runs) & 84.50 $\pm$ 0.56 & \textbf{85.49} $\pm$ \textbf{1.32} & 82.60 $\pm$ 1.12 & \textbf{86.61} $\pm$ \textbf{0.49} \\
\end{tabular}
\vspace{3pt}
\caption{Worst-class test accuracy, when the \textsc{RHO-Loss} and \textsc{Train Loss} baselines are checkpointed using their worst-class validation error during training.}
\label{tab:worst class results worst class chkp}
\end{center}
\end{table}
\vspace{-10pt}

%Average test accuracy table:
\begin{table}[h!]
\begin{center}
\begin{tabular}{lllll}
\hline
\multirow{2}{*}{\bf Dataset} &\multicolumn{4}{c}{\bf Average Test Accuracy (\%) $\pm 1$ std} 
\\

%\cline{2-7}
& \textbf{\textsc{Uniform}} & \textbf{\textsc{Train Loss}} &\textbf{\textsc{RHO-Loss}} & \textbf{\algname}
\\ \hline \\

%Just creating the table as a place holder
CIFAR10 (10 runs) & 85.09 $\pm$ 0.52 & 87.74 $\pm$ 0.50 & \textbf{89.43} $\pm$ \textbf{0.57} &  \textbf{90.02} $\pm$ \textbf{0.44} \\
CINIC10 (10 runs) & 79.57 $\pm$ 0.75 & 78.21 $\pm$ 0.57 & \textbf{81.28} $\pm$ \textbf{0.54} & \textbf{81.68} $\pm$ \textbf{0.47} \\
Clothing1M (5 runs) & 69.60 $\pm$ 0.85 &
69.46 $\pm$ 0.43 & 70.63 $\pm$ 0.87 & \textbf{72.69} $\pm$ \textbf{0.42}\\
MNLI (5 runs) & 78.85 $\pm$ 0.38 & 78.50 $\pm$ 0.33 & \textbf{80.50} $\pm$ \textbf{0.45} & \textbf{80.28} $\pm$ \textbf{0.33} \\
QQP (5 runs) & 85.23 $\pm$ 0.36 & \textbf{86.24} $\pm$ \textbf{0.26} & \textbf{86.75} $\pm$ \textbf{0.37} & \textbf{86.99} $\pm$ \textbf{0.49} \\
\end{tabular}
\end{center}
\caption{Average test accuracy, when the \textsc{RHO-Loss} and \textsc{Train Loss} baselines are checkpointed using their worst-class validation error during training.}
\label{tab:average results worst class chkp}
\end{table}

\subsection{Additional Experimental Results}

In \Cref{app:clothing1M training weights} we show the per class weights for the Clothing1M dataset, whilst in \Cref{app:clipped excess loss ablation} we analyse the effect of the clipping term and provide some intuition behind its inclusion in the algorithm.

\subsubsection{Clothing1M Training Weights}
\label{app:clothing1M training weights}

\begin{wrapfigure}{R}{0.45\textwidth}
\vspace{-10pt} 
    \centering  \includegraphics[width=0.4\textwidth]{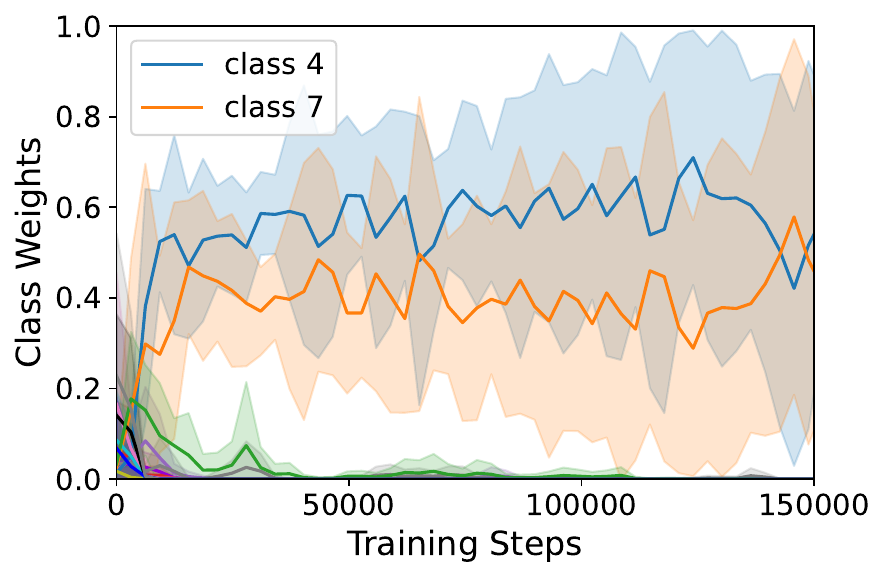}
    \vspace{-5pt}
     \caption{Clothing1M class weights \looseness=-1} 
      \label{fig:clothing1m_weights}
      \vspace{-10pt}
\end{wrapfigure}

The Clothing1M dataset is imbalanced with respect to class 4. \Cref{fig:clothing1m_weights} shows that \algname is able to consistently identify and weight the underrepresented class across model runs.

\subsubsection{Clipped Excess Loss Ablation Experiments}
\label{app:clipped excess loss ablation}

\begin{figure}[t]
     \centering
     \begin{subfigure}[t]{0.49\textwidth}
         \centering
         \includegraphics[width=\textwidth]{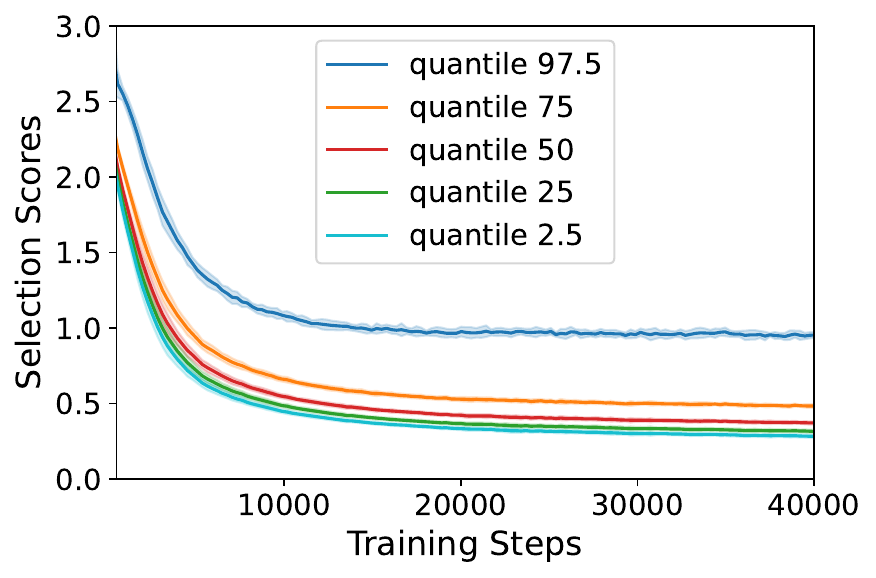}
         \caption{REDUCR Excess Loss}
         \label{fig:cinic10_selection_scores}
     \end{subfigure}
     \begin{subfigure}[t]{0.49\textwidth}
         \centering
         \includegraphics[width=\textwidth]{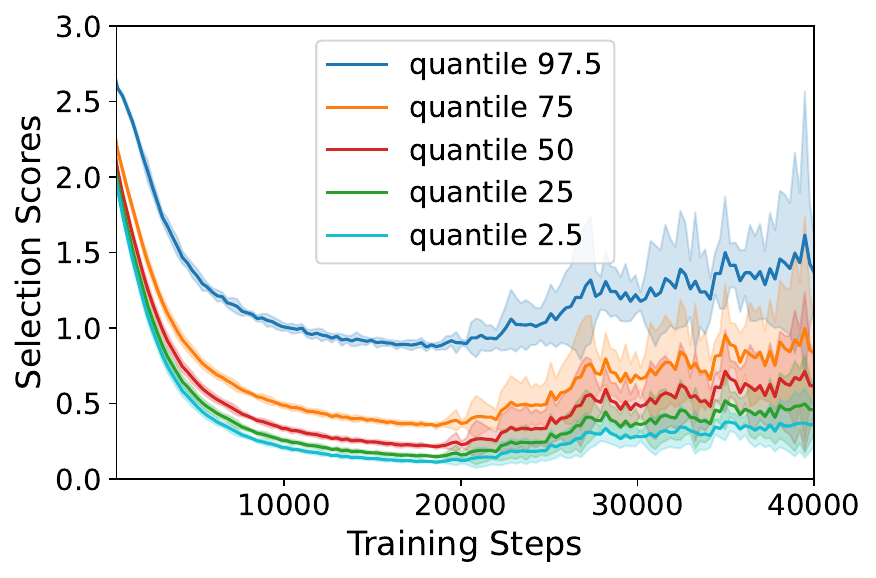}
 \caption{REDUCR Excess Loss No Clipping}   
 \label{fig:cinic10_selection_scores_non_clipped}
     \end{subfigure}
     \hfill

        \caption{The quantiles of the excess loss of points selected at each training step with (a) and without clipping (b) of the excess loss term   \looseness=-1}
        \label{fig:cinic10 selection score clip ablation}
\end{figure}

To further understand the effects of clipping in the algorithm we analyse the selection score of the selected points with and without clipping. As detailed in \Cref{sec:class holdout loss appendix} the class-holdout loss only affects the selection of points via the weights $\mathbf{w}_{t}$ at each time step, as such we record only the excess loss (the difference between the model loss and class irreducible loss). \Cref{fig:cinic10 selection score clip ablation} shows the quantiles of the weighted sum of the excess losses of points selected at each training step for the non-clipped and clipped model respectively. When the excess loss is clipped, \Cref{fig:cinic10_selection_scores} shows the selection scores smoothly decrease throughout training as the model loss improves. Without clipping the excess loss decreases smoothly at the beginning of training and then shows unstable behaviour across runs later in training. 

In practice we select multiple points per batch by selecting the points with the top k selection scores. When multiple points have the same score, points are selected at random. We note that the clipping does not reduce the excess loss of the selected points to zero where points would be selected randomly to make up the batch. 

Intuitively we posit that the clipping reduces the effect of clashing amortised class irreducible loss models in the weighted sum across the $|C|$ selection rules. The amortised class irreducible loss models are trained such that they are an expert in a specific class $c$. In some cases a model being an expert in a specific class $c'$ may result in it being a poor predictor of classes $C\setminus c'$. Even if this expert has a small weight $w_{t,c'}$ large losses may still propagate into the selection of points. Clipping the excess loss prevents a point from being down-weighted in the weighted sum of class specific selection scores by a specific class too much.\looseness=-1

\subsubsection{Alternative Solutions to Robust Data Downsampling}
\label{app:payoff}

\begin{figure}[t]
     \centering
     \begin{subfigure}[t]{0.49\textwidth}
         \centering
         \includegraphics[width=\textwidth]{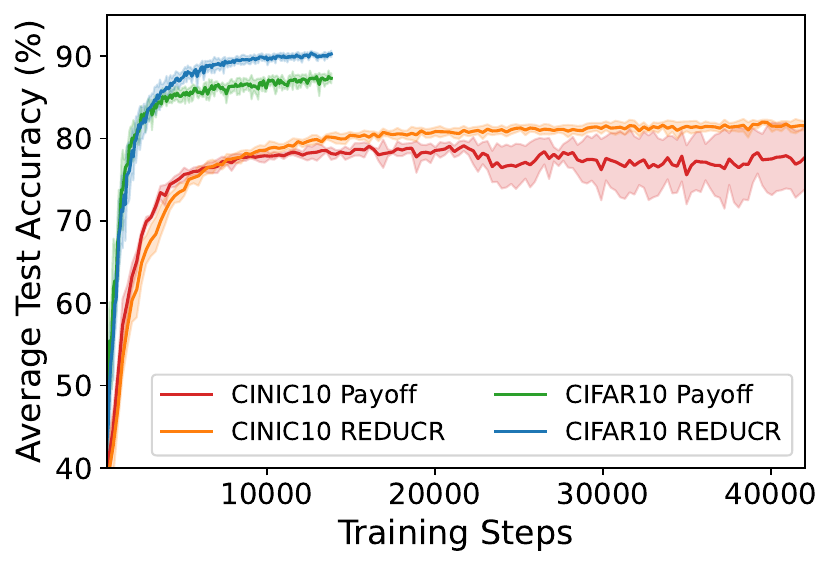}
         \caption{Average Test Accuracy}
         \label{fig:average payoff}
     \end{subfigure}
     \begin{subfigure}[t]{0.49\textwidth}
         \centering
         \includegraphics[width=\textwidth]{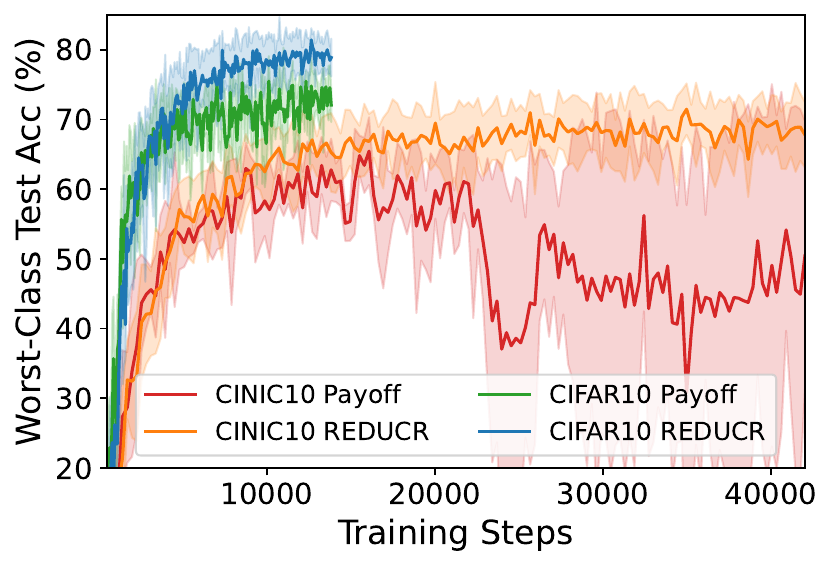}
 \caption{Worst Class Test Accuracy}   
 \label{fig:worst class payoff}
     \end{subfigure}
     \hfill
        \caption{We compare REDUCR with \Cref{alg:online batch selection payoff} in which the maximin optimisation problem approximated by solving the payoff matrix directly at each step $t$, this is labelled Payoff. \algname consistently outperforms this approach both in terms of (a) average test accuracy and (b) worst class test accuracy.         \looseness=-1}
        \label{fig:payoff}
\end{figure}

In \Cref{sec:online algorithm} we introduce an online solution to the robust data downsampling problem based upon the multiplicative weights method. In this section we empirically evaluate a variation of REDUCR where we solve the maximin optimisation problem directly. At each time step $t$ we select a small batch $b_t \subset B_t$
\begin{equation}
    b_t = \{(x,y)\} = \underset{(x,y) \in B_t}{\mathrm{argmax}} \underset{c \in [C]}{\mathrm{\min}} \mathcal{L}[y|x,\theta_t] - \mathcal{L}[y|x, \theta_{t}^{(c)}] - \mathcal{L}[\mathbf{y}_{ho}^{(c)}|\mathbf{x}_{ho}^{(c)}, \theta_t],
\end{equation}
for clarity we once again write the selection rule in terms of a single point. As $|B_t| << |\mathcal{D}|$ and both $[C]$ and $B_t$ are discrete sets it is feasible to solve the optimisation problem directly. To do this we first minimize the selection score for each datapoint $(x,y)$ with respect to a class $c$ and then select the $k$ datapoints with the greatest minimum score. The full algorithm is shown in \Cref{alg:online batch selection payoff}. We compare this approach against \algname on the CIFAR10 and CINIC10 datasets. The results can be seen in \Cref{fig:payoff} where \algname outperforms the direct optimisation approach in terms of both average test accuracy and worst class test accuracy.

\begin{algorithm}[h!]
\caption{Robust Data Downsampling Approximated by Directly Solving the Payoff Matrix}\label{alg:online batch selection payoff}
\begin{algorithmic}[1]
    \State \textbf{Input:} 
    data pool $\mathcal{D}$, holdout data $\mathcal{D}_{ho} = \bigcup_{c\in C}\mathcal{D}_{ho}^{(c)}$,
    total timesteps $T/k$, small batch size $k$

    %%FOR LOOP STARTS HERE:
    \For{$t \; \in \; [T/k]$}
    \State Sample batch $B_t$ randomly from $\mathcal{D}$
    
    \State $b_t = \underset{b \in B_t:|b|=k}{\mathrm{argmax}}\sum_{(x,y) \in b} \underset{c \in |C|}{\mathrm{\min}} \mathcal{L}[y|x,\theta_t] - \mathcal{L}[y|x, \theta_{t}^{(c)}] - \mathcal{L}[\mathbf{y}_{ho}^{(c)}|\mathbf{x}_{ho}^{(c)}, \theta_t]$ \Comment{Select points with top k min scores} 

    \State $\theta_{t+1} = \textsc{SGD}(\theta_t, b_t)$

    \EndFor
\end{algorithmic}
\end{algorithm}

\subsection{Experiments on ConvNext Architecture}

\begin{figure}[h]
     %\vspace{-1.5em}
     \centering
     \begin{subfigure}[t]{0.49\textwidth}
         \centering
         \includegraphics[width=\textwidth]{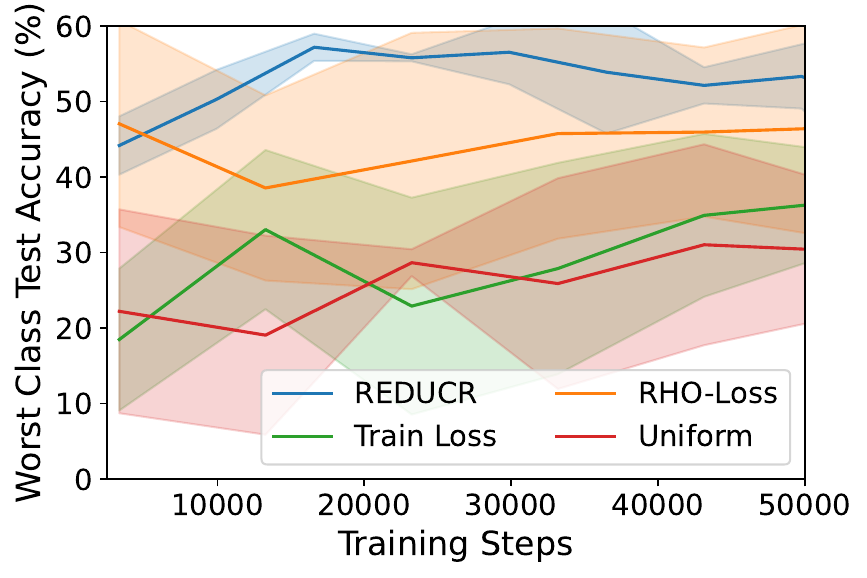}
         \caption{Worst-Class Test Accuracy (3 seeds)}
         \label{fig:clothing1m convnext worst-class test accuracy}
     \end{subfigure}
     \begin{subfigure}[t]{0.49\textwidth}
         \centering
         \includegraphics[width=\textwidth]{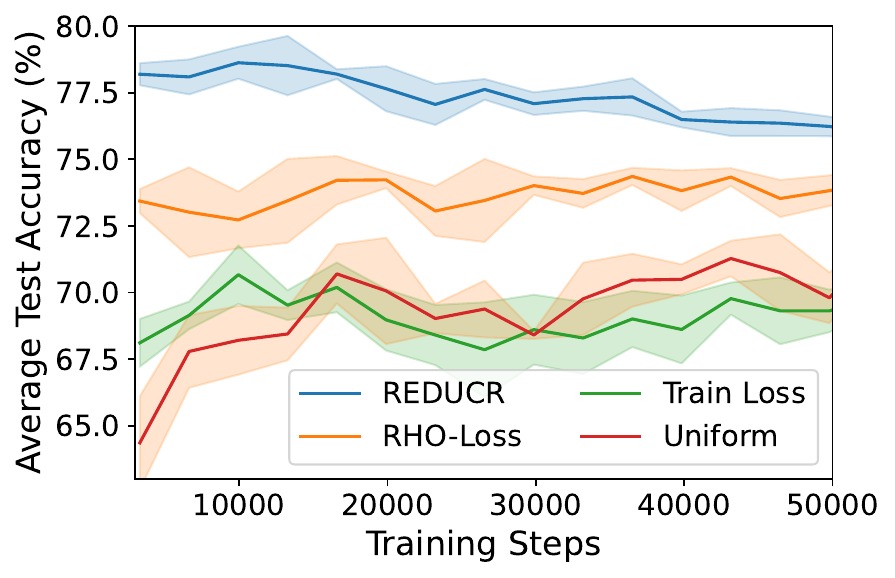}
         \caption{Average Test Accuracy (3 seeds)}
         \label{fig:clothing1m convnext average test accuracy}
     \end{subfigure}
        \caption{\subref{fig:clothing1m convnext worst-class test accuracy}) REDUCR outperforms the relevant baselines in terms of the mean worst-class accuracy using the \textit{facebook/convnext-tiny-224} ConvNext Model \citep{liu2022convnet}, we ran each experiment for 3 seeds with no hyperparameter tuning. \subref{fig:clothing1m convnext average test accuracy}) REDUCR continues to outperform all the baseline algorithms in terms of the average test accuracy on the Clothing1M dataset when training the ConvNext model. Whilst the worst-class test accuracy is noisy, the improvement REDUCR makes across multiple poorly performing classes results in a large performance difference between it and the next best performing baseline.\looseness=-1}
        \label{fig:clothing1m convnext}
\end{figure}

We repeated the Clothing1M experiments using the \texttt{ facebook/convnext-tiny-224} ConvNext model \cite{liu2022convnet} from HuggingFace, the results are shown in \Cref{fig:clothing1m convnext}. Here we note \textsc{REDUCR} maintains strong performance in terms of the average test accuracy. The mean worst-class test accuracy outperforms the other baselines, however is not statistically significant.  

\subsection{Amortised Class Irreducible Loss Models}

\begin{figure}[h!]
    \centering
    \includegraphics[width=\textwidth]{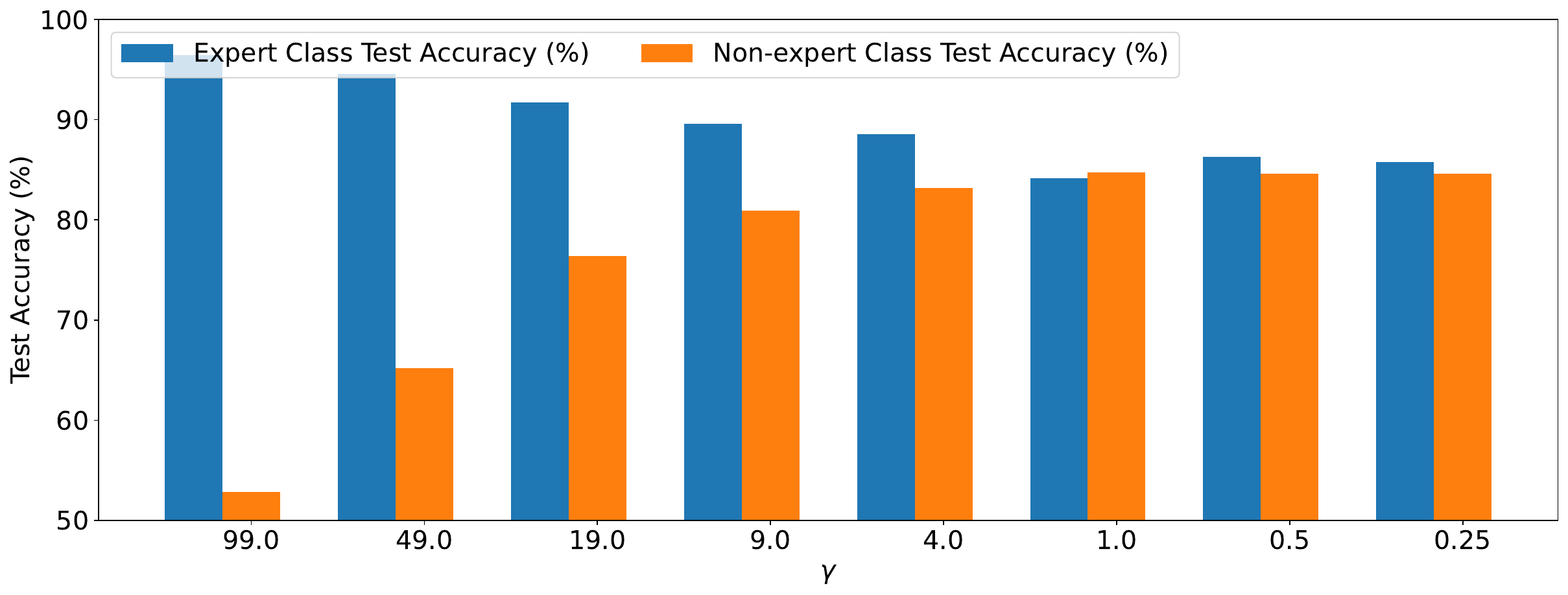}
%    \caption{CIFAR10 Class Irreducible Loss Model Test Accuracies on Expert Classes and Non-Expert Classes.}
    \caption{Class-irreducible loss model test accuracies on the expert class and non-expert classes. Class-irreducible loss models are trained using gradient weights $\gamma \in \{0.25, 0.5, 1.0, 4.0, 9.0, 19.0, 49.0, 99.0\}$.}
    \label{fig:class irred training results}
\end{figure}

In \Cref{fig:class irred training results} we compare the average expert class ($c \in C$) test accuracy and non-expert class ($c' \in C \setminus \{c\}$) test accuracy across different values of $\gamma$ for the amortised class-irreducible loss model train on CIFAR10. For the model to be an expert in one class it loses performance in the non-relevant classes. To avoid the problems described in \Cref{app:clipped excess loss ablation} we selected $\gamma = 9$ for the image datasets as the performance of the non-expert class did not suffer too much.    

\subsection{Hyperparameter Tuning}
\label{sec:hyperparameter tuning}

In this section, we test the sensitivity of \algname with respect to the hyperparameters introduced. In particular, we investigate the sensitivity of the learning rate $\eta$, used for target model training; the gradient weight $\gamma$, used for class-irreducible loss model training; the fraction of datapoints selected for target model training $|b_t|/|B_t|$, for a constant selected batch size $|b_t|$; and the frequency with which the class holdout loss term is updated during training. All experiments in this section use the CIFAR10 dataset. We use ResNet-18 target models, trained using $\eta = 10^{-4}$ with a fraction of datapoints selected of $|b_t|/|B_t| = 0.10$, and ResNet-18 class irreducible loss models trained using $\gamma = 9$ unless otherwise stated.

We find that \algname is not sensitive to the learning rate $\eta$ or the frequency of the class holdout loss term updates. We find that the performance of \algname is sensitive to the gradient weight $\gamma$ at high values. Finally, we find that \algname is not sensitive to the fraction of data points selected for target model training (referred to as the percent train) for intermediate values of percent train, though performance is poor for very low fractions and very high fractions recovers uniform selection as $b_t = B_t$ when $|b_t|/|B_t| = 1.0$ and $B_t$ is sampled uniformly from the dataset.

In summary, \algname is largely insensitive to the values of the newly introduced hyperparameters when trained on the CIFAR10 dataset. Sensitivity analyses on additional datasets are needed to increase the robustness of these findings. However, a gradient weight of $\gamma = 9$ and a percent train of $0.10$ perform well without additional hyperparameter tuning for several datasets, as shown in \Cref{tab:worst class results average checkpoint} and \Cref{tab:average results average checkpoint} which tentatively supports the robustness of these findings.

\subsubsection{Learning Rate \texorpdfstring{$\eta$}{Lg}}

\begin{figure}
     \centering
     \begin{subfigure}[t]{0.48\textwidth}
         \centering
         \includegraphics[width=\textwidth]{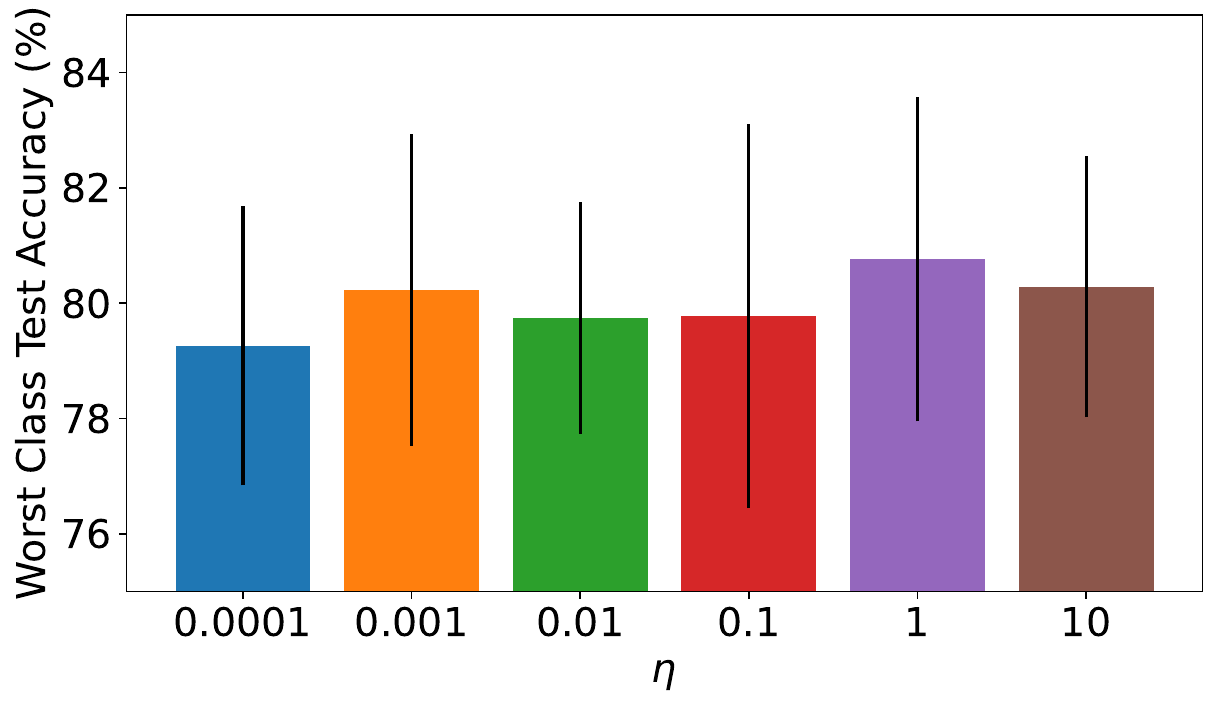}
         \caption{CIFAR10 Final Worst-Class Test Accuracy}
         \label{fig:cifar10 eta worst class final test accuracy}
     \end{subfigure}
    \hfill
     \begin{subfigure}[t]{0.48\textwidth}
         \centering
         \includegraphics[width=\textwidth]{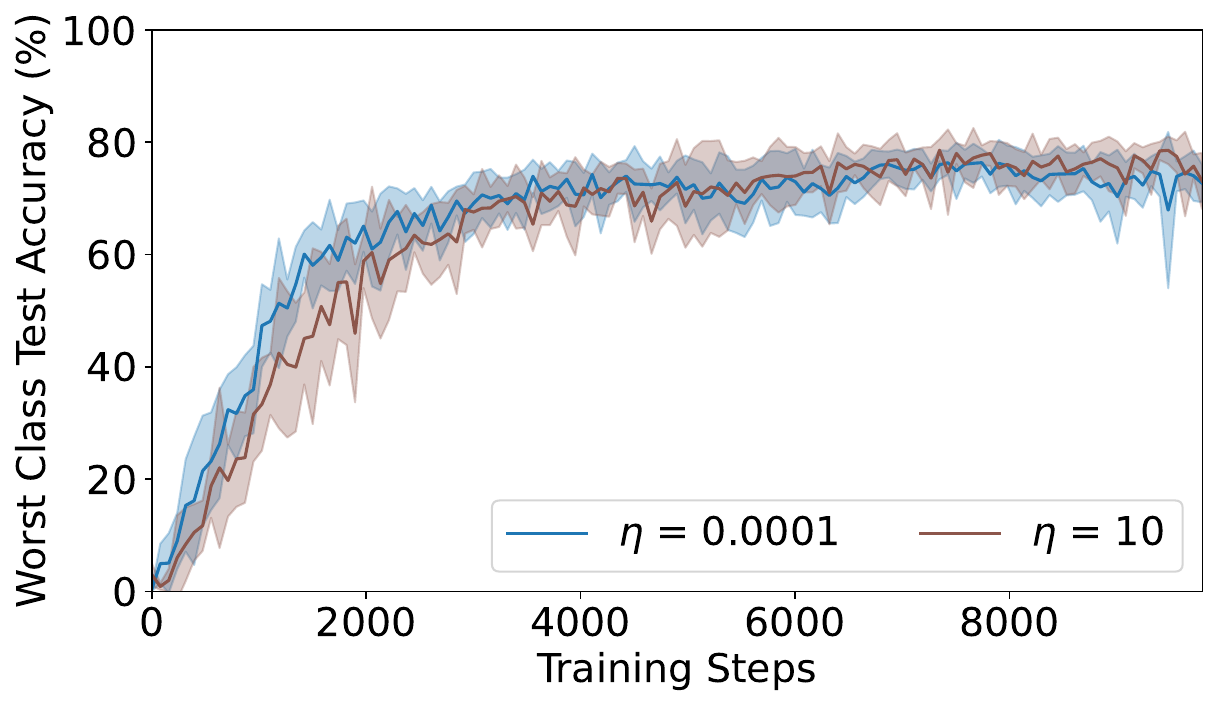}
         \caption{CIFAR10 Worst-Class Test Accuracy Curves}
         \label{fig:cifar10 eta worst class test accuracy}
     \end{subfigure}
     \begin{subfigure}[t]{0.48\textwidth}
         \centering
         \includegraphics[width=\textwidth]{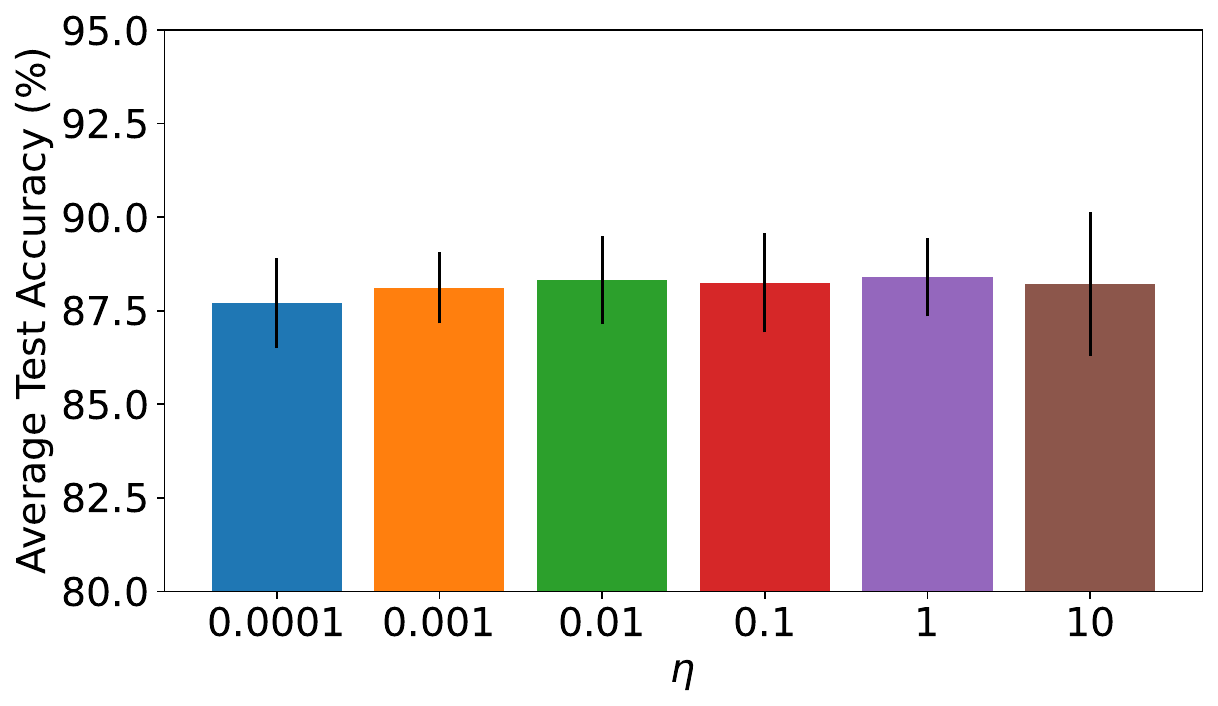}
         \caption{CIFAR10 Final Average Test Accuracy}
         \label{fig:cifar10 eta avg class final test accuracy}
     \end{subfigure}
     \hfill
     \begin{subfigure}[t]{0.48\textwidth}
         \centering
         \includegraphics[width=\textwidth]{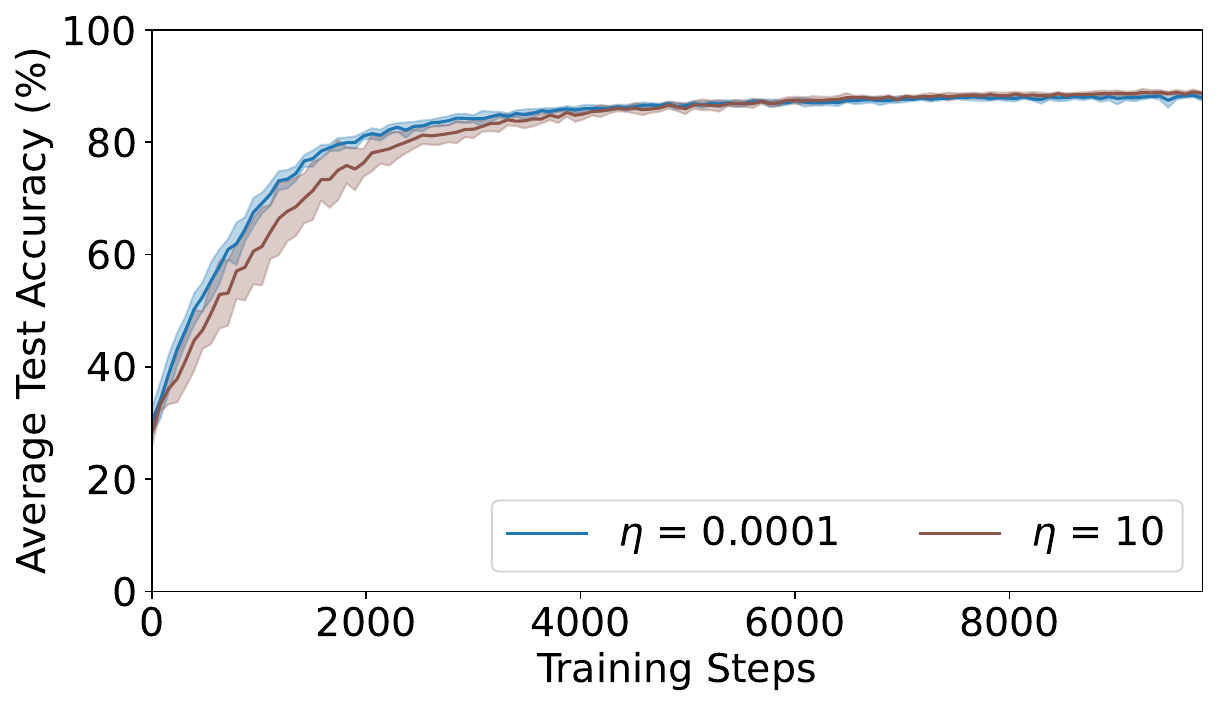}
         \caption{CIFAR10 Average Test Accuracy Curves}
         \label{fig:cifar10 eta avg class test accuracy}
     \end{subfigure}
        \caption{The final average and worst-class test accuracy are not sensitive to the value of $\eta$ on the CIFAR10 dataset. When using smaller values of $\eta$ both the average and worst class test accuracy attain higher performance at an earlier training step.}        
        \label{fig:cifar10 eta results}
\end{figure}

First, we perform a sensitivity analysis on the learning rate $\eta$ for values $\eta \in \{10^{-4}, 10^{-3}, 10^{-2}, 10^{-1}, 10^0, 10^1\}$. The experimental results, shown in \Cref{fig:cifar10 eta results},  demonstrates that smaller values of $\eta$ result in a faster improvement of the average and worst class test accuracy during training, although the final model performance is similar for all values of $\eta$ investigated. In practice, appropriately small values of $\eta$ should be used in order to reduce computational cost. Note that what constitutes an appropriately small value of $\eta$ depends on the scale of losses in a particular domain. Initial target model training runs can be done to identify a value of $\eta$ for which class weights do not prematurely concentrate on one class $\eta$.

\subsubsection{Gradient Weight \texorpdfstring{$\gamma$}{Lg}}

\begin{figure}
     \centering
     \begin{subfigure}[t]{0.45\textwidth}
         \centering
         \includegraphics[width=\textwidth]{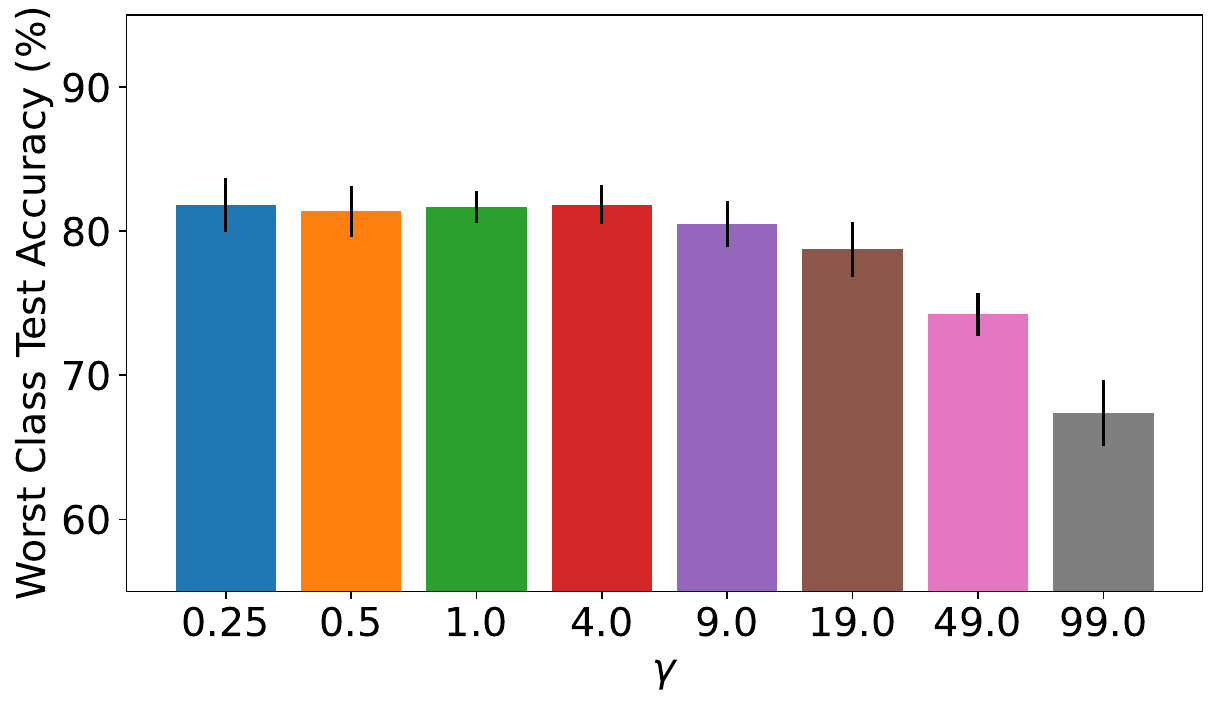}
         \caption{CIFAR10 Final Worst-Class Test Accuracy}
         \label{fig:cifar10 gamma worst class final test accuracy}
     \end{subfigure}
    \hfill
     \begin{subfigure}[t]{0.45\textwidth}
         \centering
         \includegraphics[width=\textwidth]{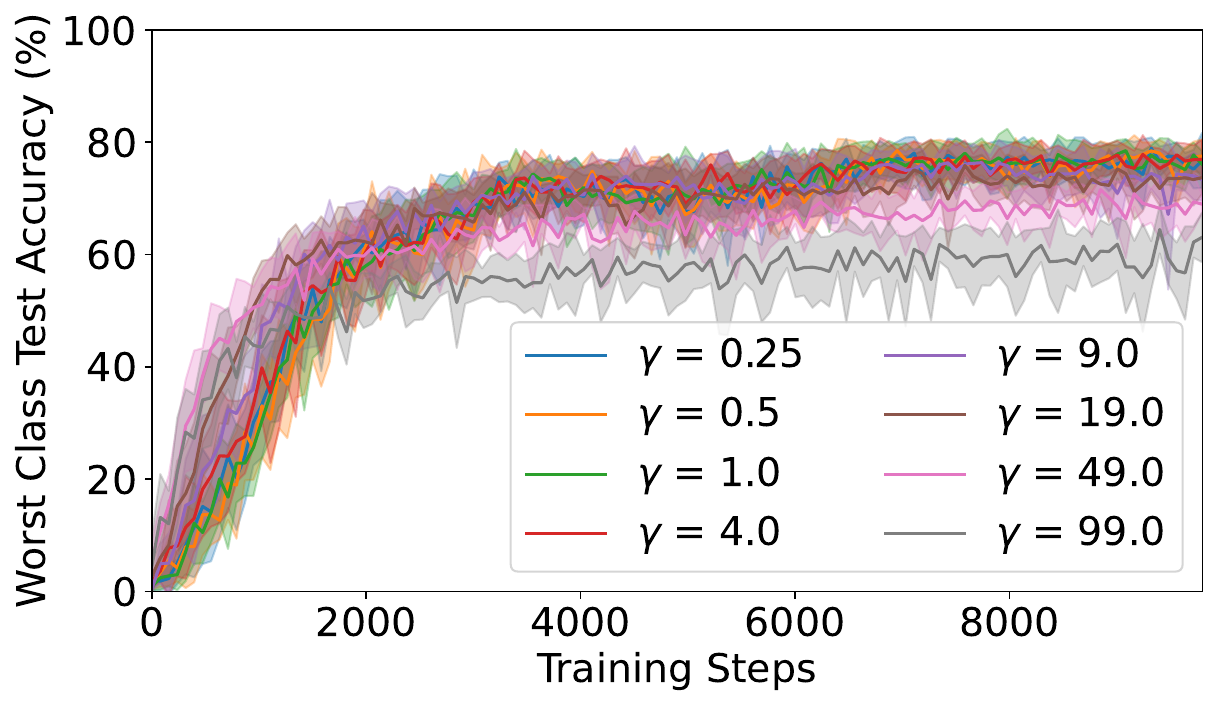}
         \caption{CIFAR10 Worst-Class Test Accuracy Curves}
         \label{fig:cifar10 gamma worst class test accuracy}
     \end{subfigure}
     \begin{subfigure}[t]{0.45\textwidth}
         \centering
         \includegraphics[width=\textwidth]{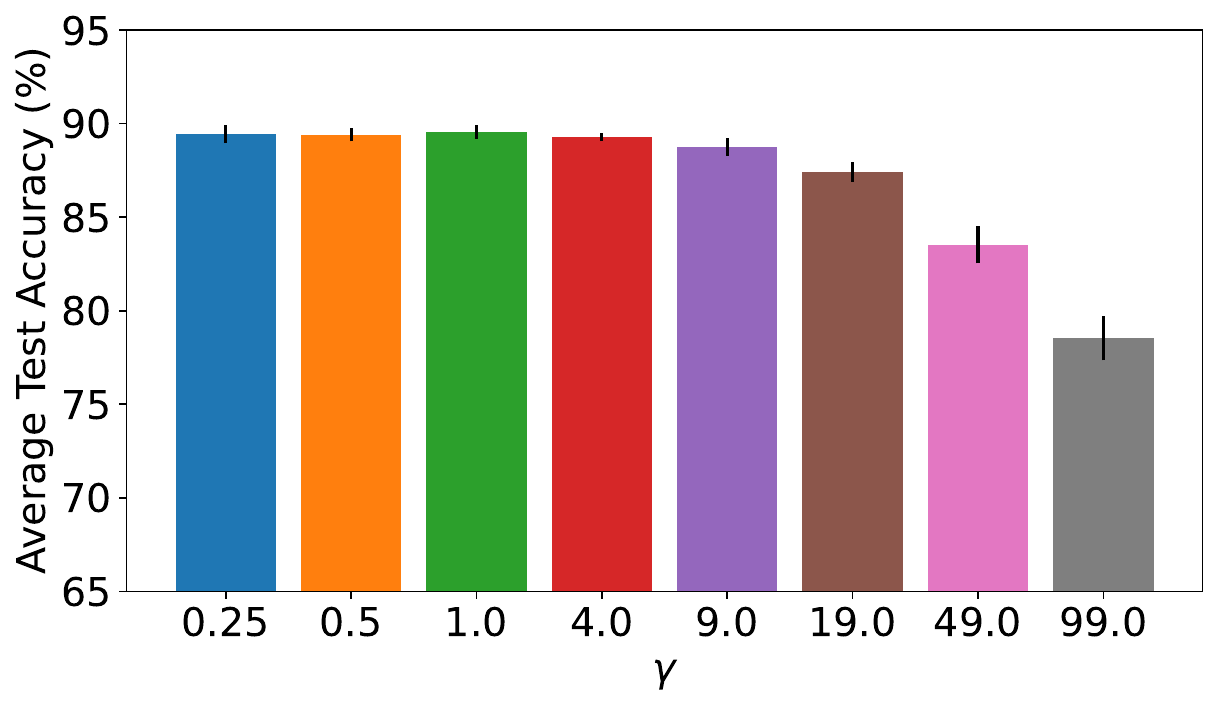}
         \caption{CIFAR10 Final Average Test Accuracy}
         \label{fig:cifar10 gamma avg class final test accuracy}
     \end{subfigure}
     \hfill
     \begin{subfigure}[t]{0.45\textwidth}
         \centering
         \includegraphics[width=\textwidth]{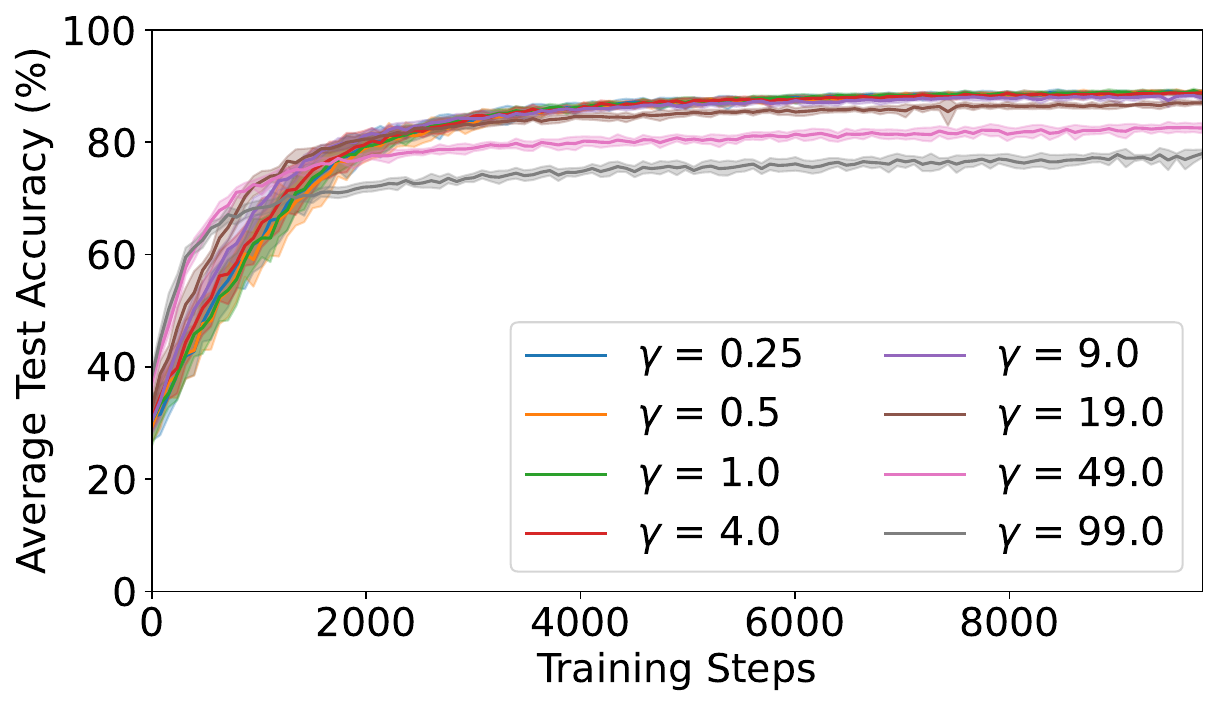}
         \caption{CIFAR10 Average Test Accuracy Curves}
         \label{fig:cifar10 gamma avg class test accuracy}
     \end{subfigure}
        \caption {\textbf{Average and worst-class test accuracy are sensitive to the value of $\gamma$ on the CIFAR10 dataset}, though this likely reflects longer convergence times for class-irreducible loss model training when using larger values of $\gamma$.}
        \label{fig:cifar10 gamma results}

\end{figure}
To investigate the sensitivity of the gradient weight $\gamma$ on the performance of \algname. We train sets of class-irreducible loss models for each $\gamma \in \{0.25, 0.5, 1.0, 4.0, 9.0, 19.0, 49.0, 99.0\}$ and train a model for each set of class-irreducible loss models. The results, shown in \Cref{fig:cifar10 gamma results}, show that values of the gradient weight above $9.0$ result in faster improvement of the average and worst class test accuracy early in training, although these models converges to a lower average and worst class test accuracy at the end of training. Higher gradient weights also increase the variance of the class-irreducible training loss, models trained with higher weights thus required a greater number of gradient descent steps to converge to a suitable class-irreducible loss model. We found class irreducible loss models trained with gradient weights $\gamma \in \{19.0, 49.0, 99.0\}$ are selected via checkpointing before the model has converged and suspect this is a contributing factor in the poor test accuracy we observed.

This finding highlights the trade-off between fast target model training, which requires a large gradient weight; fast class irreducible loss model training, which requires a smaller gradient weight; and strong final performance of the model. Final average and worst class test accuracy is similar for all models trained with gradient weights $\gamma \in \{0.25, 0.5, 1.0, 4.0, 9.0\}$.

\subsubsection{Fraction of Selected Datapoints}

\begin{figure}
     \centering
     \hspace{\fill}
     \begin{subfigure}[t]{0.45\textwidth}
         \centering
         \includegraphics[width=\textwidth]{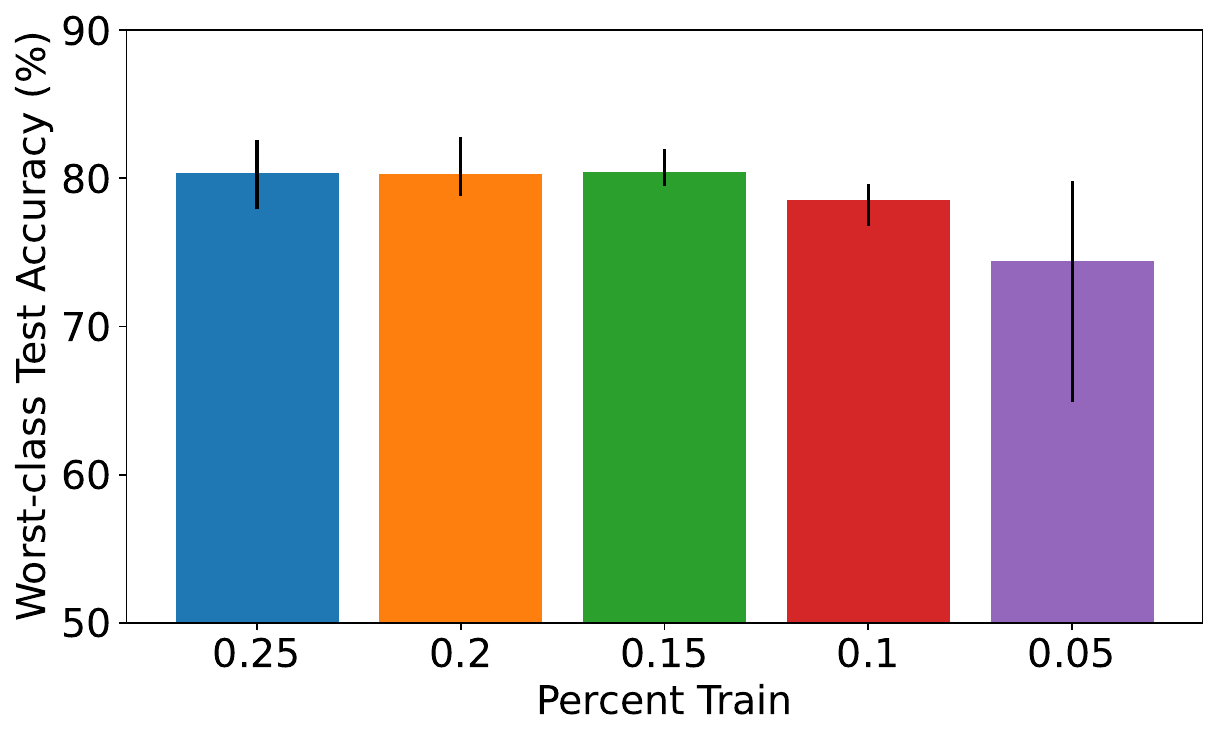}
         \caption{CIFAR10 Final Worst-Class Test Accuracy}
         \label{fig:cifar10 percent train worst class final test accuracy}
     \end{subfigure}
     \hspace{\fill}
     \begin{subfigure}[t]{0.45\textwidth}
         \centering
         \includegraphics[width=\textwidth]{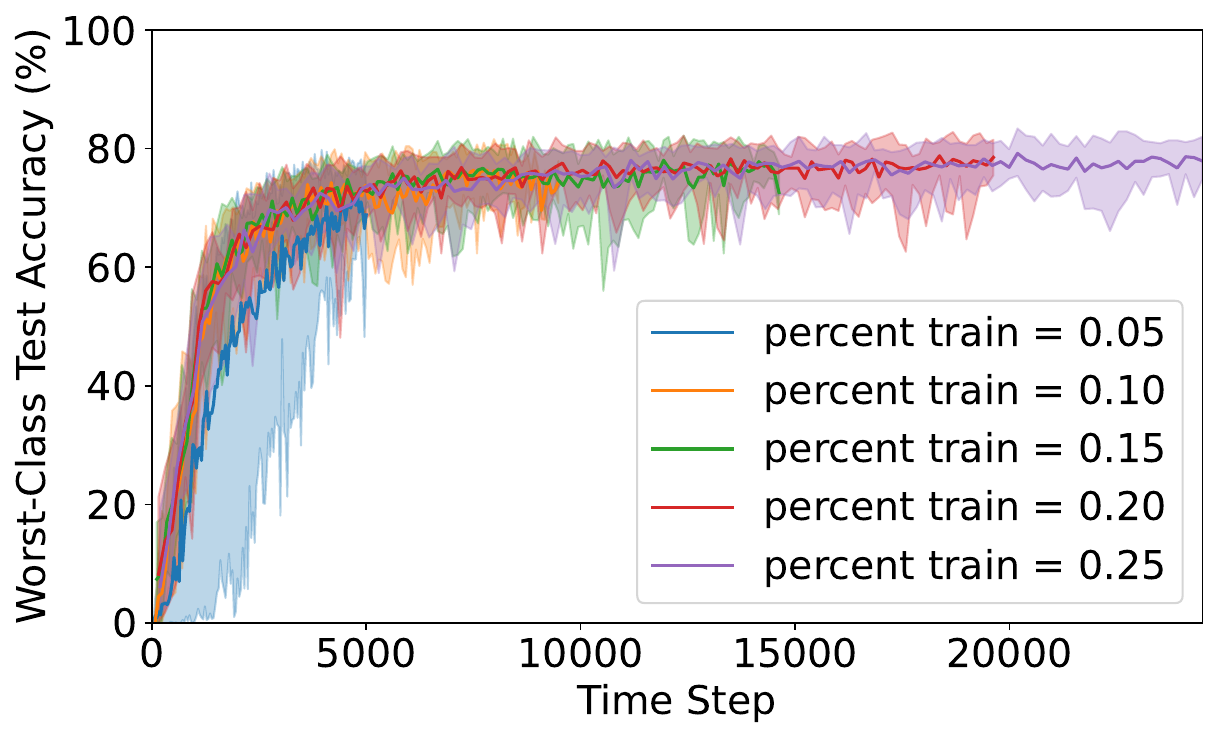}
         \caption{CIFAR10 Worst-Class Test Accuracy Curves}
         \label{fig:cifar10 percent train worst class test accuracy}
     \end{subfigure}
     \hspace{\fill}
     \begin{subfigure}[t]{0.45\textwidth}
         \centering
         \includegraphics[width=\textwidth]{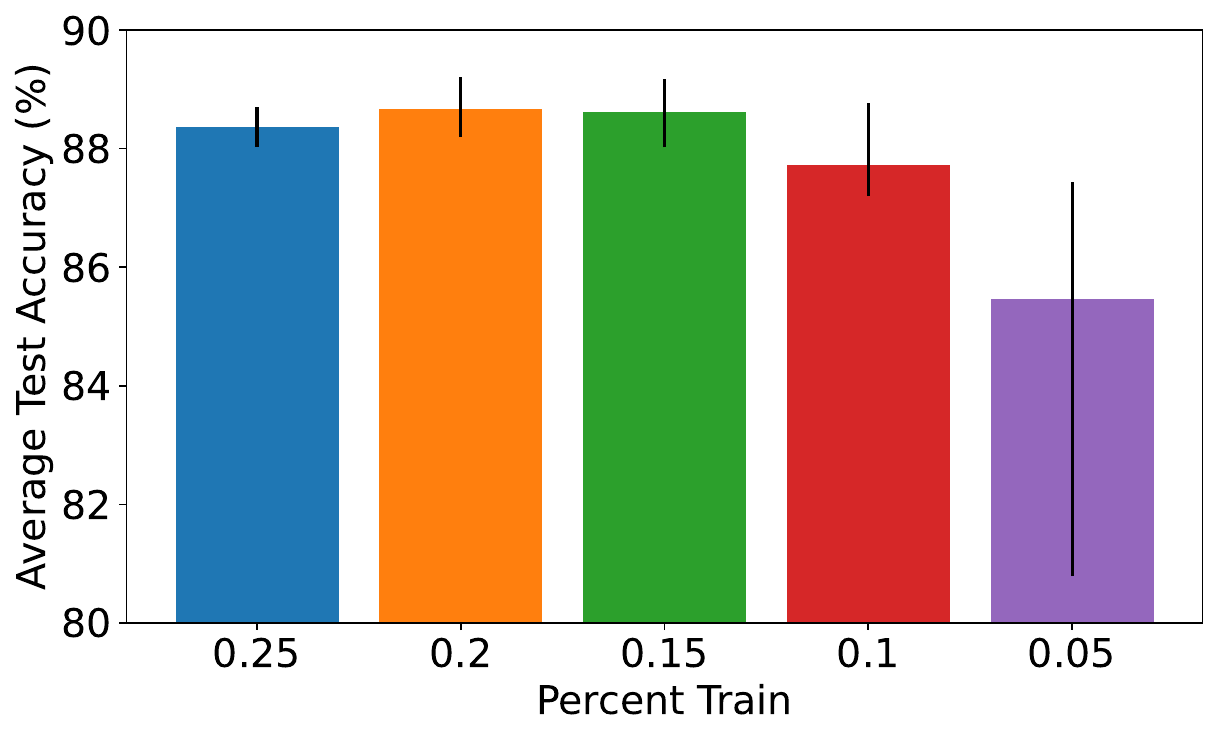}
         \caption{CIFAR10 Final Average Test Accuracy}
         \label{fig:cifar10 percent train avg class final test accuracy}
     \end{subfigure}
     \hspace{\fill}
     \begin{subfigure}[t]{0.45\textwidth}
         \centering
         \includegraphics[width=\textwidth]{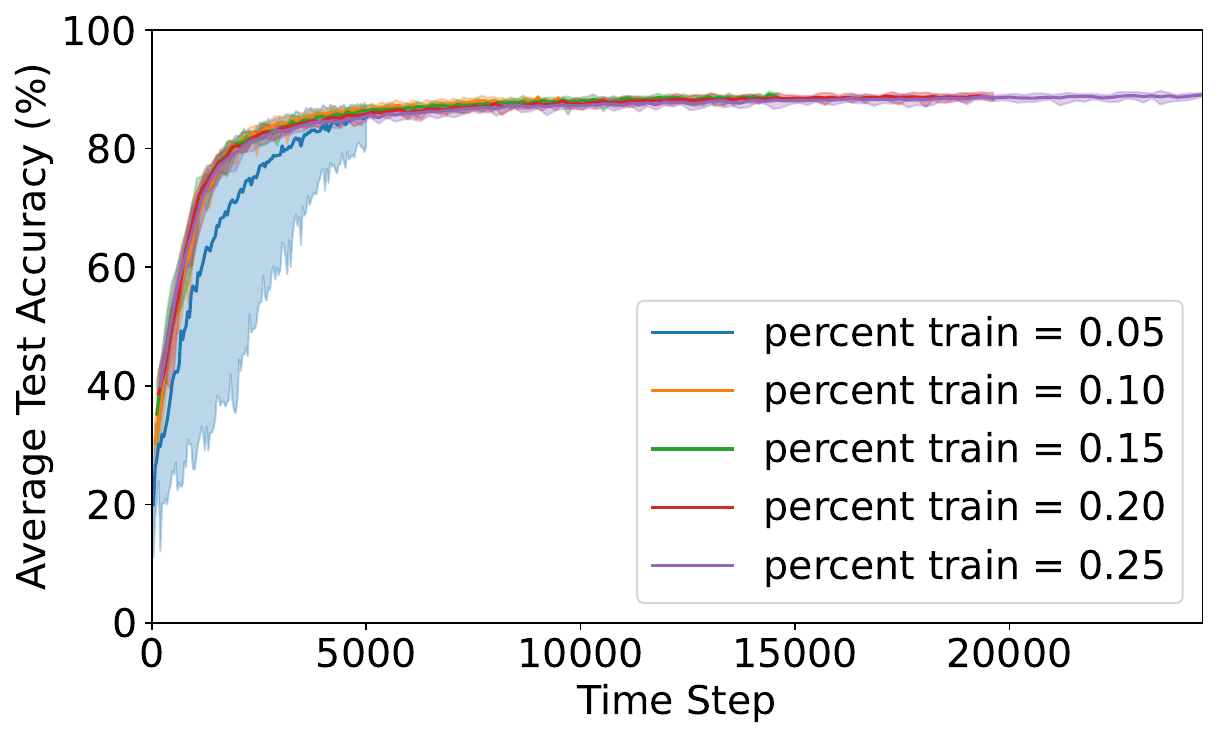}
         \caption{CIFAR10 Average Test Accuracy Curves}
         \label{fig:cifar10 percent train avg class test accuracy}
     \end{subfigure}
     \hspace{\fill}
        \caption{\algname is broadly insensitive to changes of the percent train hyperparameter, $|b_t|/|B_t|$ on the CIFAR10 dataset. The final worst class and average test accuracy is statistically similar across a variety of values of the percent train hyperparameter. For a small value of percent train $0.05$, the final worst class and average test accuracy suffers and \algname requires more gradient descent steps to match the performance of models trained with larger percent train values. The plots show the mean and standard deviation across 10 runs.}
        \label{fig:cifar10 percent train results}
\end{figure}

We perform a sensitivity analysis on the fraction of datapoints selected for target model training $|b_t|/|B_t|$, referred to as the percent train hyperparameter. We use a constant small selected batch size $|b_t|$ and vary the large batch size $|B_t|$ in order to vary the fraction of datapoints selected for target model training. In this setting, a smaller percent train allows \algname to select from a greater number of candidate datapoints at each training step, which results in the selection of datapoints with larger weighted reducible loss. Since datapoints with larger weighted reducible loss are those from which a model can learn the most \citep{mindermann2022prioritized}, we expect a smaller percent train to result in a faster improvement in target model performance.

We train target models using \algname for each percent train and large batch size pair $(|b_t|/|B_t|, |B_t|)$ in $\{(0.05, 640), (0.10, 320), (0.15, 216), (0.20, 160), (0.25, 128)\}$ and present the results in \Cref{fig:cifar10 percent train results}. We find that a percent train of $0.05$ attains lower final worst-class and average test accuracy, despite having most candidate datapoints to select from. This is surprising and is in contradiction with the intuition provided above. Furthermore, percent trains $\{0.1, 0.15, 0.2, 0.25\}$ attain similar average test accuracy at the end of training, though larger percent trains attain slightly higher worst-class test accuracy at the end of training. 

These results demonstrate that the performance of \algname is largely insensitive to the percent train  hyperparameter for a constant selected batch size. In practice, a selected batch size $|b_t|$ should first be chosen such that loss gradient estimates have a low variance, and then a large batch size $|B_t|$ should be chosen such that the percent train is an intermediate value for example $0.10$. These results also suggest that selecting datapoints with the very largest weighted reducible loss for model training may not be most appropriate for improving model performance. Instead of top-$k$ selection a more nuanced approach that accounts for the joint distribution over all points in the selected batch could be used.

\subsubsection{Frequency of Class Hold-out Loss Updating}

\begin{figure}
     \centering
     \begin{subfigure}[t]{0.40\textwidth}
         \centering
         \includegraphics[width=\textwidth]{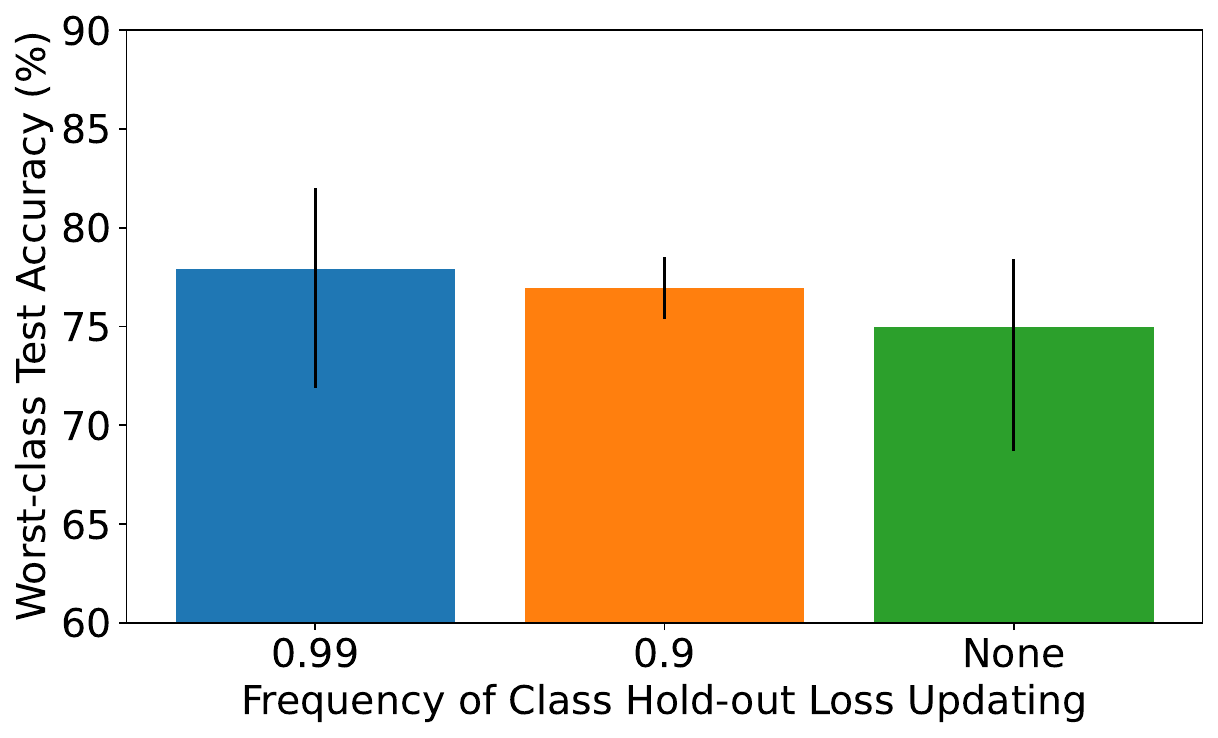}
         \caption{CIFAR10 Final Worst-Class Test Accuracy}
         \label{fig:cifar10 fast updating worst class final test accuracy}
     \end{subfigure}
    \hfill
     \begin{subfigure}[t]{0.40\textwidth}
         \centering
         \includegraphics[width=\textwidth]{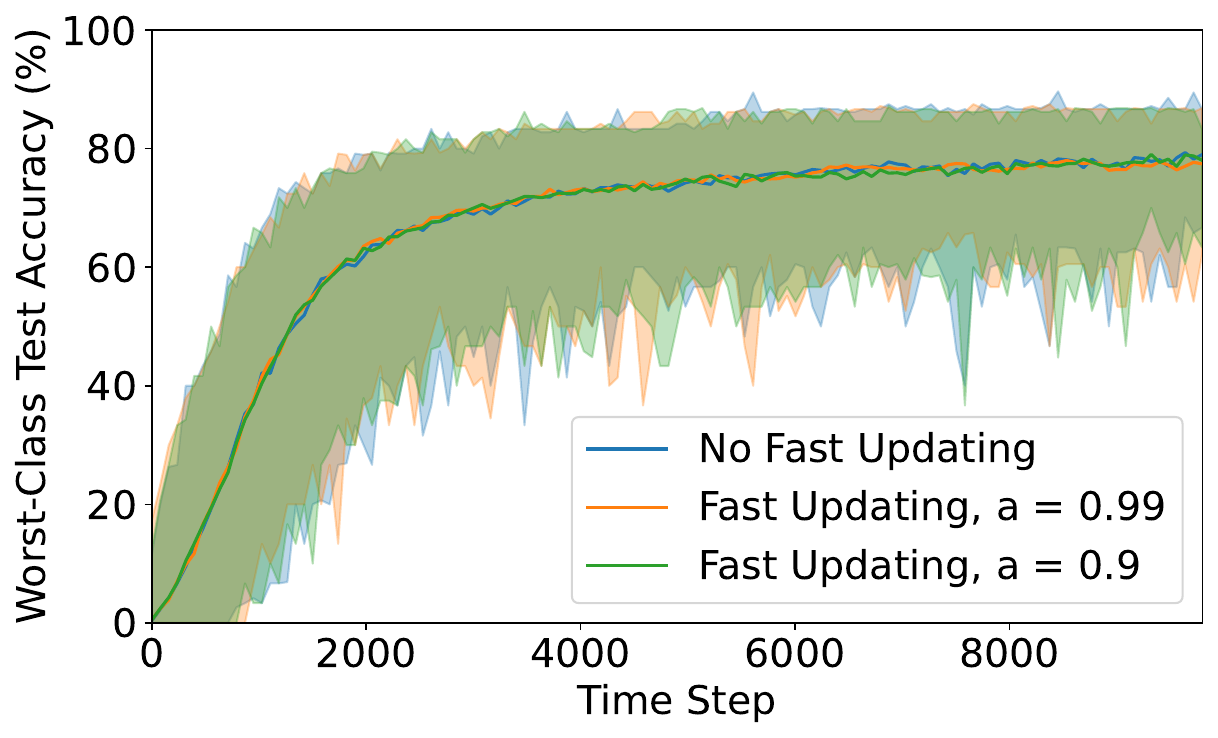}
         \caption{CIFAR10 Worst-Class Test Accuracy Curves}
         \label{fig:cifar10 fast updating worst class test accuracy}
     \end{subfigure}
     \begin{subfigure}[t]{0.40\textwidth}
         \centering
         \includegraphics[width=\textwidth]{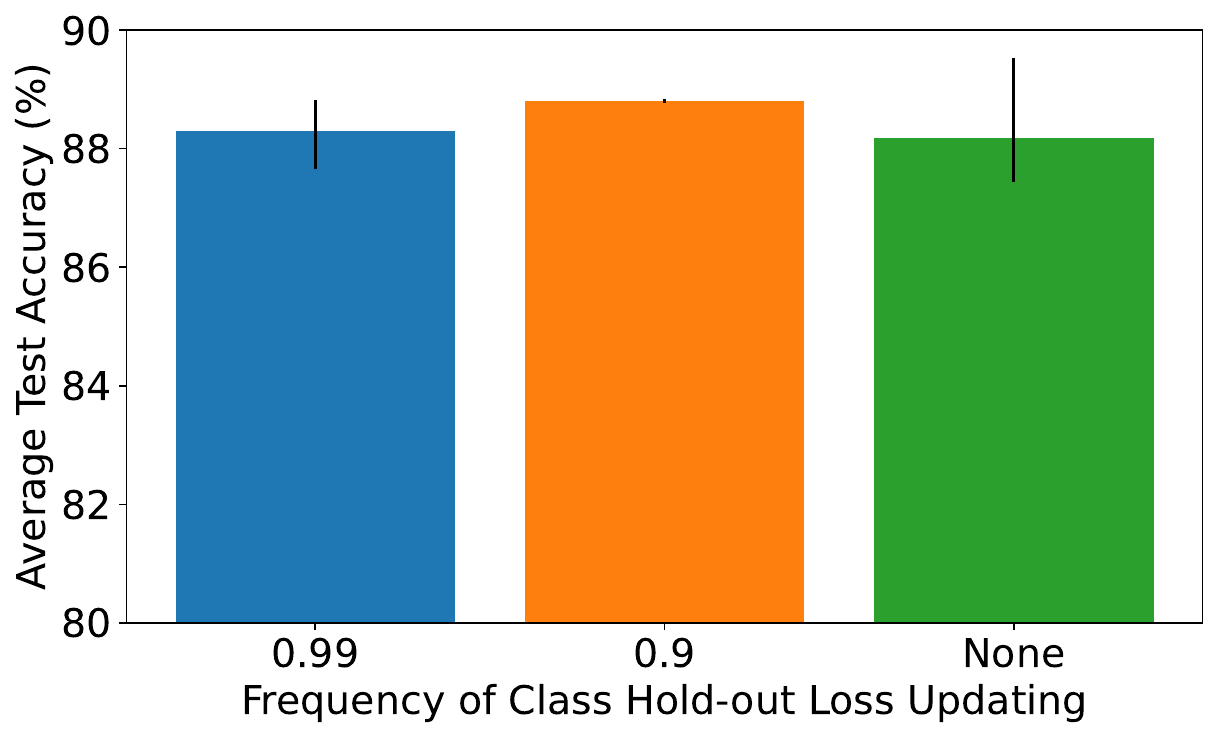}
         \caption{CIFAR10 Final Average Test Accuracy}
         \label{fig:cifar10 fast updating avg class final test accuracy}
     \end{subfigure}
     \hfill
     \begin{subfigure}[t]{0.40\textwidth}
         \centering
         \includegraphics[width=\textwidth]{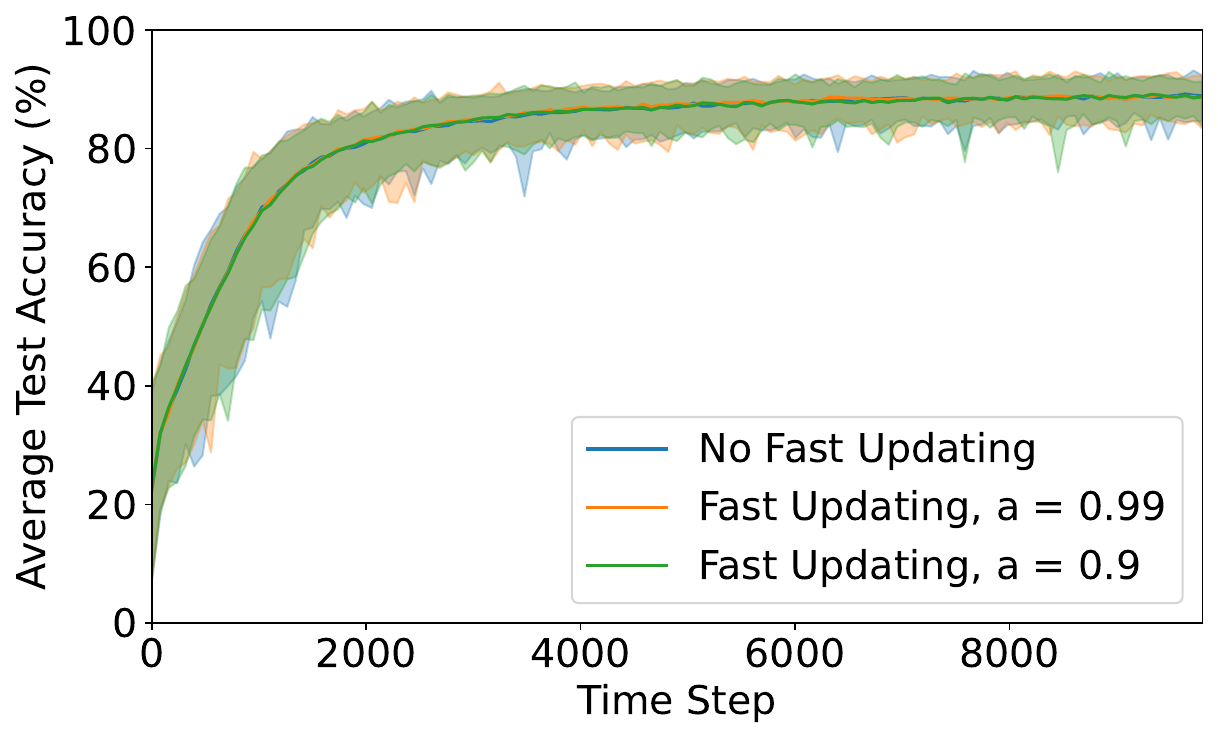}
         \caption{CIFAR10 Average Test Accuracy Curves}
         \label{fig:cifar10 fast updating avg class test accuracy}
     \end{subfigure}
        \caption{Test accuracies are not sensitive to the frequency with which class hold-out losses are updated. Plots show minimums, medians and maximum across 10 seeds.}
        \label{fig:cifar10 update frequency results}
\end{figure}

The class holdout loss term is updated at the beginning of each training epoch using the full holdout dataset. As each epoch consists of multiple gradient descent steps the actual performance of the target model on the holdout dataset will vary before the class holdout loss term is recalculated. In this section we investigate an alternative method in which the class holdout loss term is updated at every gradient descent step.    

It is computationally expensive to update the class holdout loss term using the full holdout dataset at every gradient descent step. Therefore, we propose an alternative fast updating method, which only uses a batch of holdout points to estimate the term at each timestep. The class holdout loss computed in this manner is noisy. Therefore, we use an exponentially-weighted moving average of the class holdout losses from previous timesteps to produce a smoother signal. Specifically, a batch of size $320$ is sampled uniformly at random from the holdout dataset at each training step. Losses are then computed for each datapoint in the sampled batch using the current target model. Finally, for each class $c \in [C]$,  losses of datapoints of class $c$ in the sampled batch are averaged and used to update a debiased exponentially-weighted moving average with decay parameter $a \in [0, 1]$.

We perform experiments using exponentially-weighted moving averages with decay parameters $a \in {0.9, 0.99}$ for fast updating of the class holdout loss term. The results shown in \Cref{fig:cifar10 update frequency results} demonstrate \algname is not improved in a statistically significant manner by the more complex exponentially-weighted moving average approach, both in terms of the final average and worst class test ac-curacies or the number of training timesteps required to reach these accuracies.

\subsection{Highly Imbalanced Datasets}

We also conduct experiments with 0.25\% and 0.5\% percent imbalances on classes 3 and 5. However, (class) irreducible loss models and target models only receive 6.25 and 12.5 datapoints of the imbalanced class during one training epoch (in expectation) with percent imbalances of 0.25\% and 0.5\% respectively. As a result, too few datapoints of the imbalanced class are seen during model training to achieve good performance on the imbalanced class.
\newpage
\section*{NeurIPS Paper Checklist}

%%% BEGIN INSTRUCTIONS %%%
The checklist is designed to encourage best practices for responsible machine learning research, addressing issues of reproducibility, transparency, research ethics, and societal impact. Do not remove the checklist: {\bf The papers not including the checklist will be desk rejected.} The checklist should follow the references and follow the (optional) supplemental material.  The checklist does NOT count towards the page
limit. 

Please read the checklist guidelines carefully for information on how to answer these questions. For each question in the checklist:
\begin{itemize}
    \item You should answer \answerYes{}, \answerNo{}, or \answerNA{}.
    \item \answerNA{} means either that the question is Not Applicable for that particular paper or the relevant information is Not Available.
    \item Please provide a short (1–2 sentence) justification right after your answer (even for NA). 
   % \item {\bf The papers not including the checklist will be desk rejected.}
\end{itemize}

{\bf The checklist answers are an integral part of your paper submission.} They are visible to the reviewers, area chairs, senior area chairs, and ethics reviewers. You will be asked to also include it (after eventual revisions) with the final version of your paper, and its final version will be published with the paper.

The reviewers of your paper will be asked to use the checklist as one of the factors in their evaluation. While "\answerYes{}" is generally preferable to "\answerNo{}", it is perfectly acceptable to answer "\answerNo{}" provided a proper justification is given (e.g., "error bars are not reported because it would be too computationally expensive" or "we were unable to find the license for the dataset we used"). In general, answering "\answerNo{}" or "\answerNA{}" is not grounds for rejection. While the questions are phrased in a binary way, we acknowledge that the true answer is often more nuanced, so please just use your best judgment and write a justification to elaborate. All supporting evidence can appear either in the main paper or the supplemental material, provided in appendix. If you answer \answerYes{} to a question, in the justification please point to the section(s) where related material for the question can be found.

IMPORTANT, please:
\begin{itemize}
    \item {\bf Delete this instruction block, but keep the section heading ``NeurIPS paper checklist"},
    \item  {\bf Keep the checklist subsection headings, questions/answers and guidelines below.}
    \item {\bf Do not modify the questions and only use the provided macros for your answers}.
\end{itemize}

%%% END INSTRUCTIONS %%%

\begin{enumerate}

\item {\bf Claims}
    \item[] Question: Do the main claims made in the abstract and introduction accurately reflect the paper's contributions and scope?
    \item[] Answer: \answerYes{} % Replace by \answerYes{}, \answerNo{}, or \answerNA{}.
    \item[] Justification: Our paper introduces a new methods REDUCR and claims that REDUCR improves robust data selection and sample efficiency. We provide experiments on multiple domains that support this conclusion. 
    \item[] Guidelines:
    \begin{itemize}
        \item The answer NA means that the abstract and introduction do not include the claims made in the paper.
        \item The abstract and/or introduction should clearly state the claims made, including the contributions made in the paper and important assumptions and limitations. A No or NA answer to this question will not be perceived well by the reviewers. 
        \item The claims made should match theoretical and experimental results, and reflect how much the results can be expected to generalize to other settings. 
        \item It is fine to include aspirational goals as motivation as long as it is clear that these goals are not attained by the paper. 
    \end{itemize}

\item {\bf Limitations}
    \item[] Question: Does the paper discuss the limitations of the work performed by the authors?
    \item[] Answer: \answerYes{} % Replace by \answerYes{}, \answerNo{}, or \answerNA{}.
    \item[] Justification: As per the guidelines we discuss the limitations of our work in the Conclusion, Broader Impact and Limitations section of the paper.
    \item[] Guidelines:
    \begin{itemize}
        \item The answer NA means that the paper has no limitation while the answer No means that the paper has limitations, but those are not discussed in the paper. 
        \item The authors are encouraged to create a separate "Limitations" section in their paper.
        \item The paper should point out any strong assumptions and how robust the results are to violations of these assumptions (e.g., independence assumptions, noiseless settings, model well-specification, asymptotic approximations only holding locally). The authors should reflect on how these assumptions might be violated in practice and what the implications would be.
        \item The authors should reflect on the scope of the claims made, e.g., if the approach was only tested on a few datasets or with a few runs. In general, empirical results often depend on implicit assumptions, which should be articulated.
        \item The authors should reflect on the factors that influence the performance of the approach. For example, a facial recognition algorithm may perform poorly when image resolution is low or images are taken in low lighting. Or a speech-to-text system might not be used reliably to provide closed captions for online lectures because it fails to handle technical jargon.
        \item The authors should discuss the computational efficiency of the proposed algorithms and how they scale with dataset size.
        \item If applicable, the authors should discuss possible limitations of their approach to address problems of privacy and fairness.
        \item While the authors might fear that complete honesty about limitations might be used by reviewers as grounds for rejection, a worse outcome might be that reviewers discover limitations that aren't acknowledged in the paper. The authors should use their best judgment and recognize that individual actions in favor of transparency play an important role in developing norms that preserve the integrity of the community. Reviewers will be specifically instructed to not penalize honesty concerning limitations.
    \end{itemize}

\item {\bf Theory Assumptions and Proofs}
    \item[] Question: For each theoretical result, does the paper provide the full set of assumptions and a complete (and correct) proof?
    \item[] Answer: \answerNA{} % Replace by \answerYes{}, \answerNo{}, or \answerNA{}.
    \item[] Justification: Whilst we motivate our approach from first principals we then make several approximations to develop a practical and implementable algorithm. These approximations mean that we do not have any theoretical results in the paper and support our conclusions empirically. 
    \item[] Guidelines:
    \begin{itemize}
        \item The answer NA means that the paper does not include theoretical results. 
        \item All the theorems, formulas, and proofs in the paper should be numbered and cross-referenced.
        \item All assumptions should be clearly stated or referenced in the statement of any theorems.
        \item The proofs can either appear in the main paper or the supplemental material, but if they appear in the supplemental material, the authors are encouraged to provide a short proof sketch to provide intuition. 
        \item Inversely, any informal proof provided in the core of the paper should be complemented by formal proofs provided in appendix or supplemental material.
        \item Theorems and Lemmas that the proof relies upon should be properly referenced. 
    \end{itemize}

    \item {\bf Experimental Result Reproducibility}
    \item[] Question: Does the paper fully disclose all the information needed to reproduce the main experimental results of the paper to the extent that it affects the main claims and/or conclusions of the paper (regardless of whether the code and data are provided or not)?
    \item[] Answer: \answerYes{} % Replace by \answerYes{}, \answerNo{}, or \answerNA{}.
    \item[] Justification: The paper details the full experimental setup in the main body of the paper and Appendix A.5. we also include the full code used in our experiments.
    \item[] Guidelines:
    \begin{itemize}
        \item The answer NA means that the paper does not include experiments.
        \item If the paper includes experiments, a No answer to this question will not be perceived well by the reviewers: Making the paper reproducible is important, regardless of whether the code and data are provided or not.
        \item If the contribution is a dataset and/or model, the authors should describe the steps taken to make their results reproducible or verifiable. 
        \item Depending on the contribution, reproducibility can be accomplished in various ways. For example, if the contribution is a novel architecture, describing the architecture fully might suffice, or if the contribution is a specific model and empirical evaluation, it may be necessary to either make it possible for others to replicate the model with the same dataset, or provide access to the model. In general. releasing code and data is often one good way to accomplish this, but reproducibility can also be provided via detailed instructions for how to replicate the results, access to a hosted model (e.g., in the case of a large language model), releasing of a model checkpoint, or other means that are appropriate to the research performed.
        \item While NeurIPS does not require releasing code, the conference does require all submissions to provide some reasonable avenue for reproducibility, which may depend on the nature of the contribution. For example
        \begin{enumerate}
            \item If the contribution is primarily a new algorithm, the paper should make it clear how to reproduce that algorithm.
            \item If the contribution is primarily a new model architecture, the paper should describe the architecture clearly and fully.
            \item If the contribution is a new model (e.g., a large language model), then there should either be a way to access this model for reproducing the results or a way to reproduce the model (e.g., with an open-source dataset or instructions for how to construct the dataset).
            \item We recognize that reproducibility may be tricky in some cases, in which case authors are welcome to describe the particular way they provide for reproducibility. In the case of closed-source models, it may be that access to the model is limited in some way (e.g., to registered users), but it should be possible for other researchers to have some path to reproducing or verifying the results.
        \end{enumerate}
    \end{itemize}

\item {\bf Open access to data and code}
    \item[] Question: Does the paper provide open access to the data and code, with sufficient instructions to faithfully reproduce the main experimental results, as described in supplemental material?
    \item[] Answer: \answerYes{} % Replace by \answerYes{}, \answerNo{}, or \answerNA{}.
    \item[] Justification: All the datasets used in the paper are publically available and cited where relevant. The full code for our experiments is also provided in the supplemental material. 
    \item[] Guidelines:
    \begin{itemize}
        \item The answer NA means that paper does not include experiments requiring code.
        \item Please see the NeurIPS code and data submission guidelines (\url{https://nips.cc/public/guides/CodeSubmissionPolicy}) for more details.
        \item While we encourage the release of code and data, we understand that this might not be possible, so “No” is an acceptable answer. Papers cannot be rejected simply for not including code, unless this is central to the contribution (e.g., for a new open-source benchmark).
        \item The instructions should contain the exact command and environment needed to run to reproduce the results. See the NeurIPS code and data submission guidelines (\url{https://nips.cc/public/guides/CodeSubmissionPolicy}) for more details.
        \item The authors should provide instructions on data access and preparation, including how to access the raw data, preprocessed data, intermediate data, and generated data, etc.
        \item The authors should provide scripts to reproduce all experimental results for the new proposed method and baselines. If only a subset of experiments are reproducible, they should state which ones are omitted from the script and why.
        \item At submission time, to preserve anonymity, the authors should release anonymized versions (if applicable).
        \item Providing as much information as possible in supplemental material (appended to the paper) is recommended, but including URLs to data and code is permitted.
    \end{itemize}

\item {\bf Experimental Setting/Details}
    \item[] Question: Does the paper specify all the training and test details (e.g., data splits, hyperparameters, how they were chosen, type of optimizer, etc.) necessary to understand the results?
    \item[] Answer: \answerYes{} % Replace by \answerYes{}, \answerNo{}, or \answerNA{}.
    \item[] Justification: We provide experimental setting details in Section 5 and Appendix A.
    \item[] Guidelines:
    \begin{itemize}
        \item The answer NA means that the paper does not include experiments.
        \item The experimental setting should be presented in the core of the paper to a level of detail that is necessary to appreciate the results and make sense of them.
        \item The full details can be provided either with the code, in appendix, or as supplemental material.
    \end{itemize}

\item {\bf Experiment Statistical Significance}
    \item[] Question: Does the paper report error bars suitably and correctly defined or other appropriate information about the statistical significance of the experiments?
    \item[] Answer: \answerYes{} % Replace by \answerYes{}, \answerNo{}, or \answerNA{}.
    \item[] Justification: All graphs and results in the paper are repeated over multiple runs and standard deviations are provided.
    \item[] Guidelines:
    \begin{itemize}
        \item The answer NA means that the paper does not include experiments.
        \item The authors should answer "Yes" if the results are accompanied by error bars, confidence intervals, or statistical significance tests, at least for the experiments that support the main claims of the paper.
        \item The factors of variability that the error bars are capturing should be clearly stated (for example, train/test split, initialization, random drawing of some parameter, or overall run with given experimental conditions).
        \item The method for calculating the error bars should be explained (closed form formula, call to a library function, bootstrap, etc.)
        \item The assumptions made should be given (e.g., Normally distributed errors).
        \item It should be clear whether the error bar is the standard deviation or the standard error of the mean.
        \item It is OK to report 1-sigma error bars, but one should state it. The authors should preferably report a 2-sigma error bar than state that they have a 96\% CI, if the hypothesis of Normality of errors is not verified.
        \item For asymmetric distributions, the authors should be careful not to show in tables or figures symmetric error bars that would yield results that are out of range (e.g. negative error rates).
        \item If error bars are reported in tables or plots, The authors should explain in the text how they were calculated and reference the corresponding figures or tables in the text.
    \end{itemize}

\item {\bf Experiments Compute Resources}
    \item[] Question: For each experiment, does the paper provide sufficient information on the computer resources (type of compute workers, memory, time of execution) needed to reproduce the experiments?
    \item[] Answer: \answerYes{} % Replace by \answerYes{}, \answerNo{}, or \answerNA{}.
    \item[] Justification: Compute resource details are provided in Appendix A.3.
    \item[] Guidelines:
    \begin{itemize}
        \item The answer NA means that the paper does not include experiments.
        \item The paper should indicate the type of compute workers CPU or GPU, internal cluster, or cloud provider, including relevant memory and storage.
        \item The paper should provide the amount of compute required for each of the individual experimental runs as well as estimate the total compute. 
        \item The paper should disclose whether the full research project required more compute than the experiments reported in the paper (e.g., preliminary or failed experiments that didn't make it into the paper). 
    \end{itemize}
    
\item {\bf Code Of Ethics}
    \item[] Question: Does the research conducted in the paper conform, in every respect, with the NeurIPS Code of Ethics \url{https://neurips.cc/public/EthicsGuidelines}?
    \item[] Answer: \answerYes{} % Replace by \answerYes{}, \answerNo{}, or \answerNA{}.
    \item[] Justification: Our research does not involve human subjects. In regards to data related concerns, we use dataset that are popular in academic machine learning journals and fully cite the data source, we also provide url's to the datasets as per the neurips guidelines. Our research specifically focuses upon robust methods that try to address issues with bias and discrimination by improving the performance of algorithms on these specific groups 
    \item[] Guidelines:
    \begin{itemize}
        \item The answer NA means that the authors have not reviewed the NeurIPS Code of Ethics.
        \item If the authors answer No, they should explain the special circumstances that require a deviation from the Code of Ethics.
        \item The authors should make sure to preserve anonymity (e.g., if there is a special consideration due to laws or regulations in their jurisdiction).
    \end{itemize}

\item {\bf Broader Impacts}
    \item[] Question: Does the paper discuss both potential positive societal impacts and negative societal impacts of the work performed?
    \item[] Answer:\answerYes{} % Replace by \answerYes{}, \answerNo{}, or \answerNA{}.
    \item[] Justification: 
    \item[] Guidelines:
    \begin{itemize}
        \item The answer NA means that there is no societal impact of the work performed.
        \item If the authors answer NA or No, they should explain why their work has no societal impact or why the paper does not address societal impact.
        \item Examples of negative societal impacts include potential malicious or unintended uses (e.g., disinformation, generating fake profiles, surveillance), fairness considerations (e.g., deployment of technologies that could make decisions that unfairly impact specific groups), privacy considerations, and security considerations.
        \item The conference expects that many papers will be foundational research and not tied to particular applications, let alone deployments. However, if there is a direct path to any negative applications, the authors should point it out. For example, it is legitimate to point out that an improvement in the quality of generative models could be used to generate deepfakes for disinformation. On the other hand, it is not needed to point out that a generic algorithm for optimizing neural networks could enable people to train models that generate Deepfakes faster.
        \item The authors should consider possible harms that could arise when the technology is being used as intended and functioning correctly, harms that could arise when the technology is being used as intended but gives incorrect results, and harms following from (intentional or unintentional) misuse of the technology.
        \item If there are negative societal impacts, the authors could also discuss possible mitigation strategies (e.g., gated release of models, providing defenses in addition to attacks, mechanisms for monitoring misuse, mechanisms to monitor how a system learns from feedback over time, improving the efficiency and accessibility of ML).
    \end{itemize}
    
\item {\bf Safeguards}
    \item[] Question: Does the paper describe safeguards that have been put in place for responsible release of data or models that have a high risk for misuse (e.g., pretrained language models, image generators, or scraped datasets)?
    \item[] Answer: \answerNA{} % Replace by \answerYes{}, \answerNo{}, or \answerNA{}.
    \item[] Justification: \answerNA{}
    \item[] Guidelines:
    \begin{itemize}
        \item The answer NA means that the paper poses no such risks.
        \item Released models that have a high risk for misuse or dual-use should be released with necessary safeguards to allow for controlled use of the model, for example by requiring that users adhere to usage guidelines or restrictions to access the model or implementing safety filters. 
        \item Datasets that have been scraped from the Internet could pose safety risks. The authors should describe how they avoided releasing unsafe images.
        \item We recognize that providing effective safeguards is challenging, and many papers do not require this, but we encourage authors to take this into account and make a best faith effort.
    \end{itemize}

\item {\bf Licenses for existing assets}
    \item[] Question: Are the creators or original owners of assets (e.g., code, data, models), used in the paper, properly credited and are the license and terms of use explicitly mentioned and properly respected?
    \item[] Answer: \answerYes{} % Replace by \answerYes{}, \answerNo{}, or \answerNA{}.
    \item[] Justification: All datasets are appropriately cited and URLs and data sources are provided in the Appendix A.3.
    \item[] Guidelines:
    \begin{itemize}
        \item The answer NA means that the paper does not use existing assets.
        \item The authors should cite the original paper that produced the code package or dataset.
        \item The authors should state which version of the asset is used and, if possible, include a URL.
        \item The name of the license (e.g., CC-BY 4.0) should be included for each asset.
        \item For scraped data from a particular source (e.g., website), the copyright and terms of service of that source should be provided.
        \item If assets are released, the license, copyright information, and terms of use in the package should be provided. For popular datasets, \url{paperswithcode.com/datasets} has curated licenses for some datasets. Their licensing guide can help determine the license of a dataset.
        \item For existing datasets that are re-packaged, both the original license and the license of the derived asset (if it has changed) should be provided.
        \item If this information is not available online, the authors are encouraged to reach out to the asset's creators.
    \end{itemize}

\item {\bf New Assets}
    \item[] Question: Are new assets introduced in the paper well documented and is the documentation provided alongside the assets?
    \item[] Answer: \answerNA{} % Replace by \answerYes{}, \answerNo{}, or \answerNA{}.
    \item[] Justification: The paper does not release a new dataset.
    \item[] Guidelines:
    \begin{itemize}
        \item The answer NA means that the paper does not release new assets.
        \item Researchers should communicate the details of the dataset/code/model as part of their submissions via structured templates. This includes details about training, license, limitations, etc. 
        \item The paper should discuss whether and how consent was obtained from people whose asset is used.
        \item At submission time, remember to anonymize your assets (if applicable). You can either create an anonymized URL or include an anonymized zip file.
    \end{itemize}

\item {\bf Crowdsourcing and Research with Human Subjects}
    \item[] Question: For crowdsourcing experiments and research with human subjects, does the paper include the full text of instructions given to participants and screenshots, if applicable, as well as details about compensation (if any)? 
    \item[] Answer: \answerNA{} % Replace by \answerYes{}, \answerNo{}, or \answerNA{}.
    \item[] Justification: The paper does not involve crowdsourcing nor research human subjects
    \item[] Guidelines:
    \begin{itemize}
        \item The answer NA means that the paper does not involve crowdsourcing nor research with human subjects.
        \item Including this information in the supplemental material is fine, but if the main contribution of the paper involves human subjects, then as much detail as possible should be included in the main paper. 
        \item According to the NeurIPS Code of Ethics, workers involved in data collection, curation, or other labor should be paid at least the minimum wage in the country of the data collector. 
    \end{itemize}

\item {\bf Institutional Review Board (IRB) Approvals or Equivalent for Research with Human Subjects}
    \item[] Question: Does the paper describe potential risks incurred by study participants, whether such risks were disclosed to the subjects, and whether Institutional Review Board (IRB) approvals (or an equivalent approval/review based on the requirements of your country or institution) were obtained?
    \item[] Answer: \answerNA{} % Replace by \answerYes{}, \answerNo{}, or \answerNA{}.
    \item[] Justification: The paper does not involve crowdsourcing nor research with human subjects.
    \item[] Guidelines:
    \begin{itemize}
        \item The answer NA means that the paper does not involve crowdsourcing nor research with human subjects.
        \item Depending on the country in which research is conducted, IRB approval (or equivalent) may be required for any human subjects research. If you obtained IRB approval, you should clearly state this in the paper. 
        \item We recognize that the procedures for this may vary significantly between institutions and locations, and we expect authors to adhere to the NeurIPS Code of Ethics and the guidelines for their institution. 
        \item For initial submissions, do not include any information that would break anonymity (if applicable), such as the institution conducting the review.
    \end{itemize}

\end{enumerate}

\end{document}